# Validation of an AI-based end-to-end model for prostate pathology using long-term archived routine samples

Xiaoyi Ji[1], Renata Zelic[2,3], Oskar Aspegren[2,3,4], Nita Mulliqi[5], Michelangelo Fiorentino[6], Francesca Giunchi[7], Luca Molinaro[8], Sol Erika Boman[1,2], Lorenzo Richiardi[9,10], Andreas Pettersson[3,11], Per Henrik Vincent[2,3], Martin Eklund[1], Olof Akre[2,3], Kimmo Kartasalo[5]

1. Department of Medical Epidemiology and Biostatistics, Karolinska Institutet, Stockholm, Sweden
2. Department of Molecular Medicine and Surgery, Karolinska Institutet, Stockholm, Sweden
3. Department of Pelvic Cancer, Cancer Theme, Karolinska University Hospital, Stockholm, Sweden
4. Department of Pathology and Cancer Diagnostics, Karolinska University Hospital, Stockholm, Sweden
5. Department of Medical Epidemiology and Biostatistics, SciLifeLab, Karolinska Institutet, Stockholm, Sweden
6. Department of Medical and Surgical Sciences, University of Bologna, Bologna, Italy
7. Department of Pathology, IRCCS Azienda Ospedaliero-Universitaria di Bologna, Bologna, Italy
8. Division of Pathology, AOU Città Della Salute e Della Scienza di Torino, Turin, Italy
9. Department of Medical Sciences, University of Turin, Torino, Italy
10. Cancer Epidemiology Unit, University Hospital Città della Scienza e della Salute di Torino and CPO-Piemonte, Torino, Italy
11. Clinical Epidemiology Division, Department of Medicine Solna, Karolinska Institutet, Stockholm, Sweden

**Abstract:**

Artificial intelligence (AI) is becoming a clinical tool for prostate pathology, but generalization across variations in sample preparation and preservation over prolonged time periods remains poorly understood. We evaluated *GleasonAI*, an end-to-end attention-based multiple instance learning model, on an independent validation cohort comprising 10,366 biopsy cores from 1,028 patients across 14 Swedish regions, using archival diagnostic specimens from the ProMort cohorts collected between 1998–2015. The model achieved an overall quadratic-weighted kappa of 0.86 for core-level ISUP grading, comparable to several experienced pathologists and consistent across geographic regions. Notably, performance remained stable across the 17-year collection period, demonstrating robustness to time-related variation in archival material, a property not consistently observed with foundation model-based approaches, with exploratory analysis demonstrating a significant prognostic gradient across AI-assigned grade groups for prostate cancer–specific mortality. These findings support the generalizability of the AI grading model and demonstrate the potential of pathology archives as a large-scale resource for AI development, validation, and retrospective prognostic research.

# Introduction

Prostate cancer diagnosis relies heavily on histological grading, which plays a central role in guiding treatment decisions and predicting patient outcomes[1]. The globally accepted grading system for prostate cancer is based on the Gleason score (GS) and the International Society of Urological Pathology (ISUP) grade groups, where the Gleason grading is grouped to harmonize reporting and facilitate easier communication of results[2]. Prostate cancer grading remains a task with considerable inter- and intra-observer variability. A large study in the Netherlands involving 35,258 patients revealed substantial inconsistencies in Gleason grading, both between different pathology laboratories and among pathologists within the same laboratory[3]. Over half of the labs deviated significantly from the national grading average, and 71% exhibited significant internal variation among their pathologists (based on ISUP grade 1 proportions). The challenges are further exacerbated by a global shortage of experienced urologic pathologists, as the number of prostate biopsies continues to rise. Together, these factors highlight the need for tools that can improve both the consistency and efficiency of prostate cancer grading. Artificial intelligence (AI)-assisted systems using digital histopathology images offer a promising solution, with the potential to provide standardized diagnostic support, offer nuanced and personalized prognostic approaches, and extend expert-level pathology services to resource-limited settings.

Over the past five years, AI models have made substantial progress in automating Gleason grading for prostate cancer, evolving from fully supervised methods to more scalable weakly supervised models, including the transformer-based foundation models and multiple instance learning (MIL) architectures for task-specific approaches. Among these, foundation models (FMs), large self-supervised architectures pretrained on millions of histopathology images, have demonstrated strong potential as feature extractors for downstream tasks. For example, an attention-based MIL (ABMIL) classifier[4] using features from *UNI*, a general-purpose pathology encoder pretrained on over 100 million images across 20 tissue types, achieved a Cohen's quadratically-weighted kappa (QWK) of 0.946 for ISUP grading in a large-scale prostate cancer dataset, PANDA[5,6]. While these foundation models can offer generalization and rapid adaptability, recent work by Mulliqi et al. introduced a fully end-to-end ABMIL model specifically trained for prostate cancer Gleason grading[7]. This task-specific model was developed and validated using over 100,000 biopsy cores from 15 centers across 11 countries and was benchmarked against pipelines built on leading foundation models. The end-to-end MIL model matched, and occasionally exceeded, the performance of the foundation model-based approaches for ISUP grading on both internal and external test sets. Importantly, extensive domain-focused training led to fewer clinically significant misclassifications and improved robustness across different scanners and institutions.

Despite promising performance in research settings, rigorous external validation is essential before AI models can be deployed clinically. Validation using only homogeneous internal data (i.e. data from the same source as that used for model training) may lead to overlooking of batch effects and biases that may have been inadvertently picked up by models during training. For example, biases tied to specific staining protocols, scanner characteristics, or lab workflows can lead to spurious associations and inflated performance metrics[8,9]. Studies have shown that even state-of-the-art foundation models retain site-specific signatures in their learned features, indicating persistent vulnerability to confounding technical variations[10]. When applied to external cohorts, AI models frequently experience a performance drop, especially when confronted with unseen histological artifacts or differences in tissue processing[11,12]. To ensure generalizability, validation on fully independent datasets from multiple institutions is crucial. Such evaluation can reveal whether models have truly learned disease-relevant morphologic features rather than lab-specific signatures. Accordingly, recent literature highlights the importance of large, diverse, and multi-institutional evaluations to ensure the fairness and robustness of AI systems prior to clinical integration[13].

Building on these considerations, we conducted an independent evaluation of the model by Mulliqi et al.[7], hereafter referred to as *GleasonAI*, using a fully external dataset entirely independent of the model's development and previous validation data. We leveraged the unique characteristics of the ProMort I and II datasets, derived from the National Prostate Cancer Register of Sweden (NPCR), which provided hematoxylin and eosin (H&E) stained diagnostic specimens from different municipalities across the whole of Sweden, spanning 1998-2015[14,15]. These historical slides were digitized using modern scanning technology (3DHistech Pannoramic 250 Flash II) and re-annotated by expert pathologists on a virtual microscopy platform according to contemporary ISUP grading standards, ensuring consistency with current diagnostic practices and alignment with the grading standards used to train the AI model. This validation study was implemented following a pre-specified protocol[16], adhering STARD-AI reporting guidelines (**Extended Table 9**), with the study design and statistical analysis plan established prior to model validation, ensuring methodological transparency and rigour.

The extensive temporal span of these two datasets presented an opportunity to address a critical but underexplored question in digital pathology: can AI models reliably analyze archived tissue specimens that have undergone extended storage? Historical slides are invaluable for research into prognostic models requiring long-term patient follow-up, yet present unique challenges including potential color fading, tissue degradation, and variations in historical processing protocols that differ substantially from contemporary standards[17,18]. While batch effects and technical variations are known to impact AI performance, the specific effect of temporal factors on AI-based grading remains uncharacterized. Demonstrating robust performance across decades of archived specimens would support the feasibility of using vast institutional biobanks for both AI development and retrospective outcome studies, particularly where long-term clinical outcomes are essential, such as in the case of prostate cancer. By assessing *GleasonAI* on this distinct cohort, we aim to evaluate the model's generalizability across diverse laboratory settings, inter-observer variation and, in particular, temporal variations in tissue processing—a critical step towards harvesting the full value of archived sample collections to build AI models for diseases associated with long follow-up times.

# Results

## Dataset characteristics

ProMort I comprised 1,710 matched case-control pairs of patients with low- to intermediate-risk prostate cancer[14], where cases were individuals who died of prostate cancer and controls were individuals who did not, with 290 patients (146 cases, 144 controls) contributing digitized biopsy material after quality assessment. ProMort II included 500 matched pairs of cases and controls with non-metastatic prostate cancer[15], yielding 747 patients (373 cases, 374 controls) whose specimens were digitized and had at least one cancer-containing core of sufficient quality for review. Detailed inclusion and exclusion criteria are presented in the CONSORT diagrams in the study protocol[16] (Figures 1–2). Due to varying diagnostic practices and grading guidelines over time, whole slide images (WSIs) belonging to these subjects were re-graded following the 2014 ISUP guidelines[2] and 2016 WHO Classification of Tumours of the Urinary System and Male Genital Organs by three pathologists at core level, with each chosen biopsy core individually delineated and assigned a Gleason score.

In the present analysis, evaluation was restricted to biopsy cores annotated by the reference pathologist (F.G.), who provided a consistent grading standard across the majority of the data (99.9% of ProMort I and 96.1% of ProMort II). The AI-based segmentation module successfully identified tissue in all WSIs, except for three cores (0.03%) from ProMort II in which no specimen was detected and which consequently were excluded from further analysis. After restricting to cores with reference standard grading available and excluding segmentation failures, the final validation dataset comprised 10,366 cores from 1,028 patients distributed across 14 geographic regions from 11 counties. Given the large sample size from Skåne county, this region was stratified at the municipal level to enable more granular geographic comparisons, while all other regions represent county-level data. (**Figure 1a**). The grade distribution of the datasets for validation **(Figure 1b)** is consistent with the respective inclusion criteria of each ProMort study.

## Cancer detection performance

The tissue segmentation module demonstrated robust performance, successfully detecting tissue in 99.97% of cores (10,363/10,366). Notably, manual review revealed that two of these were annotation errors where non-tissue regions had been incorrectly marked by the pathologist, while only one represented a genuinely missed small tissue fragment (**Extended Figure 1**), suggesting this AI-based segmentation could also serve as a quality control mechanism in clinical workflows.

The AI system demonstrated excellent discrimination between benign and malignant biopsy cores across both validation datasets, achieving AUC values of 0.990 (95% CI: 0.986–0.993) in ProMort I and 0.980 (95% CI: 0.977–0.983) in ProMort II (**Extended Figure 2a**). Calibration curves (**Extended Figure 2b**) showed systematic overestimation of cancer probability, particularly at higher predicted values. This phenomenon is consistent with the model's cancer probability derivation, which is designed to favor sensitivity over specificity (see Methods). As the model output is a categorical ISUP grade assignment, and the benign/malignant classification does not depend on a chosen probability threshold, the calibration of the underlying probability therefore does not directly affect the grading performance evaluated in this study.

Sensitivity and specificity were 96.9% (885/913 malignant cores correctly identified) and 93.2% (1100/1180 benign cores correctly identified) in ProMort I, and 96.2% (3098/3220) and 92.1% (4652/5053) in ProMort II. Performance remained consistent across geographic regions, with sensitivity ranging from 95–99% and specificity from 92–96% across the 5 regions in ProMort I, and slightly lower but still relatively stable performance across all 14 regions in ProMort II, demonstrating robust generalization across different laboratory settings and patient populations (**Figure 2a**). Corresponding overall and region-specific confusion matrices are provided in **Extended Figure 3**.

## Gleason grading performance

The AI system demonstrated strong agreement with the reference pathologist in grading prostate cancer across both validation datasets. The QWK for ISUP grading across all cores (including benign) was 0.86 [0.84–0.88] in ProMort I and 0.86 [0.85–0.87] in ProMort II; when restricted to malignant cores, QWK was 0.71 [0.66–0.75] and 0.68 [0.56–0.71], respectively. For Gleason score classification, corresponding values were 0.81 [0.78–0.84] and 0.83 [0.81–0.84] for all cores, and 0.68 [0.64–0.72] and 0.67 [0.65–0.69] for malignant cores. Examination of the normalized confusion matrices (**Figure 3a**) revealed that agreement was highest for benign cores and ISUP grade 5, while intermediate grades showed patterns consistent with established areas of diagnostic difficulty. ISUP 1 cores showed a tendency toward AI predictions of equal or one grade higher, ISUP 2 and 4 cores were predominantly assigned within one adjacent grade, and ISUP 3 cores displayed the broadest dispersion of AI predictions across ISUP 2 to 5, reflecting the well-documented challenge of distinguishing Gleason pattern 3 from pattern 4 and differentiating 3+4 from 4+3.

### Subgroup analysis: geographic variation

Regional subgroup analyses showed consistent grading performance across diverse geographic regions in both validation datasets (**Figure 2b**), with ISUP grade QWK values ranging from 0.79–0.89 in ProMort I and 0.80-0.92 in ProMort II. Similarly consistent results were observed for Gleason scores, with QWK values ranging between 0.71–0.88 across both datasets. When restricted to malignant cores, agreement values exhibited greater variability across regions, with ISUP QWK ranging from 0.57–0.81 in ProMort I and 0.52–0.78 in ProMort II. The malignant-only analysis exhibited wider confidence intervals, particularly in regions with smaller cancer sample sizes, though the performance levels were within the expected ranges observed in the AI model's prior external validations[7,19], and also within the commonly reported range of inter-pathologist agreement (approximately 0.5–0.8)[20]. Detailed evaluation metrics (linearly weighted kappa (LWK), QWK, ordinal C-index) are provided in **Extended Table 2–3**.

## Sensitivity analysis: sample collection time variation

To assess the impact of temporal factors on model performance, cores from the combined ProMort I and II datasets were grouped by sample collection date into equal temporal intervals, with ISUP grade distributions balanced across intervals through stratified resampling to minimize confounding from grade distribution differences (**Figure 4a**). Analysis revealed stable grading performance across sample collection periods spanning from 1998 to 2015. For ISUP grading on all cores, QWK values ranged from 0.83 to 0.87 across five temporal subsets, with a subtle trend toward improved performance in more recent samples (2012–2015: QWK 0.87 [0.81–0.92]) compared to older specimens (1998–2001: QWK 0.83 [0.81–0.86]). Similar temporal stability was observed for Gleason scoring (QWK 0.78–0.82).

To control for potential geographic confounding, the analysis was repeated using only cores from Skåne-Malmö (**Figure 4b**). This single-source analysis confirmed performance robustness, where ISUP QWK values were 0.88, 0.90, and 0.86 for samples collected during 1998–2003, 2003–2008, and 2008–2013, respectively. The consistent performance across samples collected over 15 years demonstrates the model's robustness to temporal variation, and its resilience to potential changes in specimen characteristics over time. Detailed evaluation metrics for all temporal subsets are presented in **Extended Table 4–5**.

## Sensitivity analysis: inter-observer agreement

To contextualize the AI model's grading performance within the inherent subjectivity of prostate cancer grading, inter-observer agreement was evaluated on subsets of cores with multiple pathologist annotations. In ProMort I, 548 malignant cores with Gleason score disagreement between the reference pathologist (F.G.) and an independent reviewer (L.M.) were adjudicated by a third pathologist (M.F.), while in ProMort II, 347 randomly sampled cores received independent review from three pathologists (F.G., M.F., Os.A.), who represented three different institutions in Italy and Sweden. Details on the composition and selection of these two subsets are provided in the study protocol[16].

For each human observer, agreement was calculated as the average QWK against the other two pathologists, and for the AI model, as the average QWK against all three pathologists. In the ProMort I subset, inter-pathologist concordances for ISUP grading were 0.57, 0.64 and 0.67, with the AI model achieving 0.62, ranking third among the four observers (**Figure 4c**). The ProMort II subset demonstrated notably higher overall agreement levels but comparable patterns, with ISUP grade concordance ranging from 0.90 to 0.94 among pathologists and the AI model achieving 0.90 (**Figure 4d**). Similar results were observed for Gleason score, with the AI model performing within the inter-pathologist range. Overall, its agreement across both subsets was comparable to that observed among experienced uropathologists, underscoring that the model performs on par with human experts despite the inherent subjectivity of prostate cancer grading. Detailed evaluation metrics for all inter-observer agreements are presented in **Extended Table 6**.

### Exploratory analysis: prognostic association

To preliminarily illustrate the feasibility of prognostic analyses based on archived material, we performed an exploratory patient-level survival analysis using AI-assigned ISUP grades and Gleason scores. For this analysis, all available biopsy slides from each patient were combined into a model input, allowing the AI system to directly infer a patient-level grade. The analysis was limited to the AI-based grading, because directly comparable patient-level reference grading was not available, and any type of rule-based aggregation of pathologists' core-level grades would not necessarily reflect actual clinical patient-level grading and could introduce artefactual differences, precluding a fair comparison. Patients with slides deemed inadequate during digitization review were excluded, with 965 patients remaining for evaluation.

Using ISUP grade 1 (GS 3+3) as reference, a monotonic increase in hazard ratios was observed from ISUP 3 through ISUP 5 (**Figure 3b**), with all grade groups ≥3 showing statistically significant associations with prostate cancer mortality ($p < 0.001$). ISUP 2 showed a non-significant elevation in risk (HR 1.27 [0.90–1.80], p=0.176), consistent with the known prognostic similarity between ISUP 1 and 2 in low-risk disease. The AI-predicted benign group (ISUP 0, n=58 with 18 events) showed no significant difference from ISUP 1 (HR 1.22 [0.72–2.07], p=0.460); this group likely includes false-negative cases among patients with confirmed prostate cancer, which is reflected by the wide confidence interval. Within ISUP 5, Gleason score sub-groups showed an internal gradient (GS 4+5: HR 4.72; GS 5+4: HR 6.32; GS 5+5: HR 6.84), suggesting the AI model distinguishes prognostically distinct Gleason patterns. Notably, the AI-assigned grades also captured the prognostic distinction between Gleason scores 3+4 (ISUP 2) and 4+3 (ISUP 3), with patients graded as 4+3 showing nearly twice the risk of prostate cancer–specific mortality (HR 1.98 [1.34–2.91], $p=5.7\times10^{-4}$). Detailed analysis values are presented in **Extended Table 7**.

## Comparison with foundation models

As an exploratory analysis, *GleasonAI* was compared against two general-purpose pathology foundation models adapted for prostate cancer grading (**Figure 5**). *UNI* utilizes a ViT-L/16 architecture[5] and *Virchow2* a ViT-H/14 architecture[21], both pre-trained using the DINOv2 self-supervised framework[22]. For direct performance comparison, both foundation models were used as frozen feature extractors, with identical aggregation and classification modules trained on the same dataset as *GleasonAI* for Gleason grading tasks, as described in Mulliqi et al.[7]

For cancer detection, the foundation models generally showed higher sensitivity but lower specificity than *GleasonAI* across most regions (**Figure 5a-b**). In ProMort I, the foundation models achieved higher sensitivity in four of five regions, generally at the expense of specificity; in ProMort II, they exceeded *GleasonAI* in sensitivity in thirteen of fourteen regions, with specificity reduced in most cases. For Gleason score, the *GleasonAI* achieved superior agreement in the majority of regions, achieving the highest QWK in 4 of 5 regions in ProMort I and 7 of 14 regions in ProMort II, ranking second elsewhere (**Figure 5c**). To compare temporal robustness across model architectures, the same stratified temporal analysis was applied to the two foundation model-based pipelines (**Figure 5d**). While *GleasonAI* maintained stable agreement across all five temporal intervals, both UNI and Virchow2 showed a pattern of declining performance on older specimens, with progressively lower QWK values for both Gleason score and ISUP grade agreement for samples collected further from the present. Overall, the end-to-end model demonstrated more balanced performance between sensitivity and specificity for cancer detection, while maintaining competitive or superior grading agreement across geographic regions, and exhibited greater temporal stability relative to the foundation models.

# Discussion

In this study, comprehensive external validation was performed for an end-to-end deep learning system for prostate cancer detection and Gleason grading, demonstrating robust performance across 10,366 biopsy cores from routine clinical practice among 1,028 patients spanning 14 Swedish healthcare regions. The AI model exhibited strong performance for both cancer detection and grading tasks, maintaining consistency across diverse geographic regions, different laboratory practices, and samples collected during a 17-year period.

The *GleasonAI* system demonstrated strong discriminatory ability for cancer detection, and exhibited a consistent automated operating point favoring sensitivity over specificity across geographic regions, indicating a lower decision threshold for cancer detection relative to the reference pathologists—preferentially flagging potential malignancy in ambiguous cases. This sensitivity-favoring pattern reflects the design choices of *GleasonAI*, which was constructed to act as a safety net for the pathologists to not miss positive cases, with the rationale that the reporting pathologists would increase specificity by downgrading false positives. This characteristic aligns with observations from the original validation study, suggesting consistent optimization rather than a dataset-specific artifact.

The AI system demonstrated strong agreement with the reference pathologist and remained consistent across all 14 geographic regions despite differences in local laboratory practices. A stratified resampling analysis balancing ISUP grade distributions confirmed that these regional patterns reflect genuine generalization rather than confounding by case-mix differences (**Extended Figure 4**). Inter-observer analysis provided important context for interpreting performance: in ProMort I, which consists of diagnostically challenging cases with initial reviewer disagreement, the AI model operated within the range of human variability; in ProMort II, the model achieved intermediate inter-observer agreement relative to the three pathologists, suggesting potential utility as a consistent reference across varying diagnostic practices. Beyond aggregate agreement metrics, to further characterize model behavior, we examined patch-level cancer probability and Gleason pattern predictions in representative cores, visualized as probability heatmaps across tissue regions (**Figure 6**). Although trained only with slide-level labels, the model learned meaningful spatial localization, producing heatmaps that aligned with tissue morphology rather than technical image features. Concordant examples illustrate that the model highlights the same architectural patterns identified by expert reviewers, while discordant cases reveal that differences tend to arise in borderline or diagnostically ambiguous regions. Together, these findings demonstrate that the model's predictions reflect clinically relevant morphology, with uncertainties and disagreements that parallel human interpretation, supporting its potential to provide consistent second opinions in clinical practice.

A notable finding was the model's stable performance across samples collected over 17 years (1998–2015), demonstrating robustness to temporal variation in tissue processing and preparation methods. Performance remained unaffected by sample collection year or storage duration, and this long-term stability has important implications for AI applications in digital pathology. To our knowledge, this study represents the first validation of AI assisted cancer diagnosis systems based on histopathology images across an extensive temporal span, where sample age is specifically considered. The model's ability to operate effectively on archival material indicates that digitized historical slides can be directly repurposed for both model development and evaluation, without requiring retrieval of tissue blocks for re-sectioning or re-staining. This is practically important, as recent AI studies using archival prostate cancer trial material have relied on retrospective block retrieval and fresh slide preparation[23], likely reflecting concern that ageing of archived material could compromise downstream digital analysis. At the same time, the temporal robustness is particularly relevant for outcome-based research, where long follow-up periods necessitate the use of archival specimens, which are often collected years or decades before clinical endpoints are observed. The exploratory survival analysis demonstrated the practical utility of this capability: despite training solely for diagnostic grading without outcome supervision, the model's grade assignments showed a clear prognostic gradient, which further provided evidence that the AI model has learned biologically meaningful diagnostically and prognostically relevant morphological features rather than superficial correlations.

Comparison with foundation models revealed stable operating-point differences across data subsets: both foundation models showed higher sensitivity and lower specificity for cancer detection than the end-to-end model. For Gleason grading, the end-to-end model achieved marginally higher agreement with the reference pathologist in the majority of subgroups, also consistent with the original validation findings. Beyond replicating these prior observations, the temporal analysis further revealed that while all three models performed comparably on recent specimens, the foundation model-based pipelines showed declining concordance on older samples, whereas *GleasonAI* maintained stable performance across the full 17-year span. Together, these findings suggest that while foundation models provide strong general-purpose representations, task-specific end-to-end optimization may offer advantages both for capturing the fine-grained morphological distinctions required for prostate cancer grading and for robustness to temporal variations in tissue characteristics, potentially because joint optimization of the feature encoder allows learning of representations more invariant to archival artifacts and historical processing differences.

There are a few limitations worth consideration. First, the validation cohort was limited to Swedish patients, predominantly of European ancestry, necessitating further validations in more diverse populations to ensure generalization[19]. Second, during preprocessing, we only extracted patches that overlapped with pathologist-annotated core areas. This approach did not permit us to fully examine the segmentation model performance on processing potential artifacts, background tissue, or ambiguous regions that would be encountered in automated clinical deployment, consequently our reported segmentation success rate might overestimate the real-world performance. Last, it is worth mentioning that the validation dataset exhibited heterogeneous annotation granularity due to differences in core delineation: ProMort I and a subset of ProMort II (n=54) included manually delineated cores, while the remaining ProMort II samples employed automated segmentation with minimal manual adjustment, potentially resulting in partial core fragments (detailed in the study protocol[16]). This finer granularity represents a more stringent evaluation, as the model is required to correctly classify smaller tissue regions with less contextual information, which indicates that our reported performance may underestimate the model's capability in standard whole-slide evaluation.

In conclusion, beyond validating this specific AI system, our findings highlighted the underutilized potential of pathology archives accumulated over decades of clinical practice. The demonstrated temporal robustness, with diagnostic and prognostic value preserved in specimens up to almost two decades old, indicated that historical slides can be digitized and repurposed for AI development and validation without requiring tissue reprocessing. This substantially expands available data resources and supports the feasibility of retrospective prognostic studies relying on archival material. As digital pathology infrastructure continues to mature, such archives represent a valuable resource with potential to accelerate both methodological development and clinical translation of AI-assisted diagnostics.

# Methods

## Study materials

### ProMort I

ProMort I is a nested case-control study derived from the National Prostate Cancer Register of Sweden (NPCR), originally designed to develop prognostic models for patients of low- to intermediate-risk prostate cancer[14]. From approximately 57,952 eligible individuals diagnosed between January 1, 1998, and December 31, 2011, 1,710 pairs of cases, defined as patients who died from prostate cancer, and controls (who were alive when the matched case died and matched on year and hospital of diagnosis), were included in the original study.

After the digitization of the ProMort I diagnostic specimens between November 2015 and February 2016, samples from 313 patients from Örebro and Skåne counties were pathologically re-reviewed in order to confirm low- to intermediate-risk classification and establish interobserver concordance. Three genitourinary pathologists (F.G., L.M., M.F.) were involved in the pathological review, annotating independently following a structured annotation protocol.

During the digitization of the prostate biopsies and the re-review process, we performed quality control by excluding whole slide images (WSIs) of tissue not stained with hematoxylin and eosin (H&E). The final ProMort I Gleason grading validation dataset comprised 2,096 cores from 1,780 slides from 290 patients (146 cases and 144 controls) for the AI system validation analysis.

## ProMort II

ProMort II represents a nested case-control sample of patients diagnosed with non-metastatic prostate cancer between January 1, 1998, and December 31, 2015, derived from the NPCR[15]. Cases and controls, defined as in ProMort I, were selected from 11 of 21 Swedish counties, with 500 matched case-control pairs randomly selected from this population (controls were matched on year and county of diagnosis and were alive at the time of the corresponding case's event).

Slide digitization was performed from May 2017 to January 2018 for all 1,000 subjects, successfully yielding digitized biopsies for 404 cases and 426 controls. Of these, 60 patients from Örebro and Värmland counties were randomly selected to evaluate the interchangeability of standard light microscopy and virtual microscopy systems[24], with three pathologists (F.G., M.F., Os.A.) conducting independent annotations. The remaining 770 subjects underwent pathological review by a single pathologist (F.G. or M.F.)[15].

Similarly as for ProMort I, we conducted quality control procedures excluding WSIs of tissue not stained with H&E and slides rejected during the pathological review process but inadvertently retained in the dataset. The final ProMort II Gleason grading validation dataset consisted of two parts: a case-control subsample with 8,242 cores from 3,862 slides (693 subjects, 351 cases and 342 controls), and a software validation subsample with 374 cores from 322 slides (54 subjects, 22 cases and 32 controls).

All prostate biopsies were digitized using a Pannoramic 250 Flash II scanner (3DHistech Ltd., Budapest, Hungary) at 40X magnification (0.19 µm/pixel resolution)[14,15,24], with WSIs stored in MRXS format. Digital pathological review was performed using a virtual microscopy system developed in collaboration between Centre for Advanced Studies, Research and Development in Sardinia (CRS4), Pula, Italy and the ProMort study utilizing the OMERO platform[25]. All pathologists' reviews included initial slide evaluation (such as tissue identification and staining assessment) and quality control for annotation suitability, Gleason grading, and precise spatial delineation of individual cores (individual cylindrical tissue specimens from needle biopsies) within each WSI, with coordinate data stored in JSON format. Both ProMort validation datasets represent external validation cohorts, with samples collected from clinical sites and laboratories independent of those used for the AI systems' training[26]. For CONSORT diagrams with detailed inclusion and exclusion criteria, see Figure 1-2 in the study protocol[16]. For information on the reference standard protocols, illustrating distribution of pathologist reviews and inter-reviewer overlap across subsamples, see Table 3 in the study protocol[16].

## AI systems

### WSI pre-processing

All WSIs underwent automated tissue detection using a UNet++ architecture[27] with ResNet-18[28] encoder pretrained using semi-weakly supervised learning[29]. The process began with the extraction of 512×512 pixel patches at 8.0 μm/px resolution across the entire WSI with 128-pixel overlap[30]. These patches underwent pixel-wise tissue segmentation, which was subsequently assembled into a comprehensive binary tissue mask for each WSI. Then high-resolution patches of 256×256 pixels were extracted at 1.0 μm/px resolution from tissue-positive regions, requiring a minimum of 10% tissue content per patch, with 50% overlap (128 pixels per stride) to enhance diagnostic accuracy. The patches were downsampled to 1.0 μm/px using Lanczos resampling from the nearest higher resolution level in the image pyramid.

Using the spatial coordinate annotations provided by pathologists to delineate individual biopsy cores within each WSI, extracted patches were assigned to their corresponding cores based on spatial overlap with the annotated core boundaries. This approach enabled core-level analysis by organizing patches according to their corresponding annotated cores, enabling AI evaluation consistent with the pathological reference standard. For each core with a Gleason grade annotation, all corresponding patches were stored in an individual TFRecord[31] file for computational efficiency.

### Deep learning models

**Model architecture**: The AI model for validation employed an attention-based multiple instance learning (ABMIL) architecture trained end-to-end for Gleason pattern classification, using an EfficientNet-V2-S encoder[32] to extract 1,280-dimensional feature representations from individual tissue patches, which underwent dimensionality reduction to 1,000-dimensional vectors through average pooling and a fully connected layer. These features were then processed through a gated-variant ABMIL aggregator that computes attention weights for each patch, enabling the model to focus on the most diagnostically relevant regions within each input set. The attention-weighted features were aggregated into a higher level (core, slide or patient level) representation for classification. The output layers provided probability distributions for primary and secondary Gleason patterns across four classes: benign, Gleason pattern 3, pattern 4, and pattern 5.

**Model training**: The model was trained end-to-end, i.e. all model components including the feature encoder were jointly optimized. The training dataset included 61,483 WSIs from 4,467 patients within three clinical cohorts (Capio S:t Göran Hospital, Stavanger University Hospital, and Stockholm3[33])[26], encompassing multiple laboratories, scanning platforms, and varied tissue characteristics. The model was trained using 10-fold cross-validation on data from these three clinical cohorts, resulting in an ensemble of 10 models used for Gleason grading validation. Complete training methodology is described in Mulliqi et al.[7]

**Model prediction**: Model predictions employed the ensemble of 10 models with an optimized test-time augmentation strategy. Instead of applying 3 augmentation iterations per model as in the published approach (yielding 30 predictions total)[7], we implemented a more efficient scheme using a single systematic augmentation per model. Instead of stochastically applying random augmentations to the input tiles, we applied augmentations deterministically using a set of 8 unique combinations of geometric transformations, including no transformation, horizontal/vertical flips, and 90°/270° rotations. The 8 transformations were assigned cyclically across the 10 models, such that each model processed a differently augmented set of input tiles using a single fixed transformation, yielding 10 predictions in total instead of 30, while still applying a degree of test-time augmentation coupled with model ensembling.

The ensemble produced core-level classifications, where the Gleason scores were aggregated through majority voting of the 10 model outputs, with corresponding ISUP grades derived according to standard guidelines (detailed in the study protocol[16]). A benign diagnosis was encoded as Gleason score 0+0 (ISUP grade 0). Additionally, the cancer probabilities were aggregated as the median of individual model scores, each derived from the underlying Gleason pattern probability vectors. Patch-level classifications, providing Gleason pattern probabilities for individual patches within each core, were generated by utilizing the trained feature encoder independently on individual patches, circumventing the attention aggregation mechanism to provide granular pattern probability maps for tissue visualization.

**Foundation models**: We conducted exploratory analyses to evaluate end-to-end versus transfer-learning-based foundation models alongside our primary validation. Two foundation models were included for comparative assessment: *UNI*, utilizing a ViT-L/16 architecture pre-trained with DINOv2 self-supervised learning to generate 1,024-dimensional patch embeddings[5], and *Virchow2*, employing a ViT-H/14 architecture with the same pre-training framework to produce 1,280-dimensional embeddings[21]. For both foundation models, we employed their frozen pre-trained encoders as feature extractors in a transfer learning approach, maintaining the same model inference procedures, including ABMIL aggregation and classification components, as the task-specific model. The foundation models' training and prediction procedures followed a workflow identical to the primary model, using the same training data, 10-fold cross-validation structure, and ensemble prediction approach, enabling direct performance comparison between end-to-end and transfer learning strategies.

## Statistical analysis

Performance of the AI system was evaluated using multiple metrics to assess both cancer detection and Gleason grading accuracy. For cancer diagnosis (positive/negative), we assessed agreement with the reference standard using sensitivity (true positive rate), specificity (true negative rate), and area under the receiver operating characteristic curve (AUROC). Calibration was assessed by plotting observed cancer frequencies against predicted cancer probabilities using a logistic calibration model fitted with a generalized additive model (GAM) smooth term (*cal_plot_logistic* from the R *probably* package, with *smooth=TRUE*), with 95% confidence intervals. Gleason score and ISUP grade concordance was quantified using quadratic weighted Cohen's kappa (QWK) and linear weighted Cohen's kappa (LWK), with ordinal C-index calculated to measure the model's ability to correctly rank patients according to cancer grade severity with discrete predictions. Analyses for cancer grading were conducted using two approaches: including all cores (benign and malignant) and restricting to cores with malignant diagnoses by the reference pathologist. Subgroup analysis was conducted across geographic regions using the same metrics to evaluate the generalizability of the AI system across diverse patient populations. Uncertainty around all performance estimates was quantified using 95% confidence intervals (CIs) calculated through non-parametric bootstrap resampling with 1,000 iterations.

Sensitivity analyses were conducted to evaluate performance robustness. Performance across different sample collection periods was assessed employing stratified sampling within each period to achieve comparable ISUP grade distributions, thereby minimizing the effects of grade distribution differences on temporal performance comparisons. Inter-observer agreement was evaluated using subsets of cores annotated by multiple pathologists, with the AI system included to evaluate its agreement relative to human variability. Additionally, exploratory survival analysis was performed to assess whether AI-assigned grades retained prognostic information. Cox proportional hazards regression was used to estimate hazard ratios for prostate cancer–specific mortality across AI-predicted grade groups, with ISUP 1 (Gleason 3+3) as the reference category. ISUP grade was modelled as a categorical variable. For ISUP grades containing multiple Gleason scores (ISUP 4 and 5), hazard ratios were additionally estimated at the Gleason score level. Absolute survival rates were not estimated, as the nested case-control design does not yield population-representative event rates, and the concordance statistics were likewise not reported. Annotated cores where the segmentation model failed to detect any tissue were documented and excluded from all analyses, with the failures reported separately and analyzed by an expert pathologist to determine potential causes to provide transparency regarding technical limitations of the AI system. Detailed analysis plans are provided in the study protocol[16].

## Computing hardware and software

**Programming environment and dependencies:** All AI model inference experiments were conducted using Python (v3.8.10) with PyTorch (v2.0.0, CUDA 12.2) as the deep learning framework. Pre-trained foundation model weights for *UNI* and *Virchow2* were obtained from their official HuggingFace repositories (https://huggingface.co/MahmoodLab/UNI; https://huggingface.co/paige-ai/Virchow2) and integrated using ViT implementations from the timm library (v0.9.8).

**WSI pre-processing and data handling:** WSI access was facilitated through OpenSlide[34] (v4.0.0) and openslide-python (v1.3.1), while tissue segmentation models were implemented using PyTorch's segmentation_models_pytorch library (v0.3.3)[35]. Core-level patch assignment was performed using shapely (v2.0.6), specifically utilizing Polygon.intersects functionality to determine spatial overlap between annotated core boundaries and extracted patches. Data format compatibility between TFRecord files and PyTorch[36] was achieved using DareBlopy (v0.0.5). Image augmentations were applied through Albumentations[37] (v1.3.1) and Stainlib[38] (v0.6.1), with basic image processing tasks handled by Pillow (v9.4.0) and OpenCV-python[39].

**Computing infrastructure:** All experiments were executed on a single 40 GB A100 NVIDIA GPU on the high-performance cluster Berzelius at the National Supercomputer Centre at Linköping University. Containerization was managed through Docker (v20.10.21) for local image preprocessing and Singularity/Apptainer for cluster-based model inference.

**Analysis and visualization:** Data management and statistical analyses utilized NumPy (v1.24.0), scikit-learn (v1.2.2), Pandas (v1.5.3) and lifelines (v0.30.0) in Python and probably (v1.1.1) in R, while visualization was accomplished using Matplotlib (v3.7.1) and Seaborn (v0.12.2) in Python and ggplot2 (v4.0.0).

# Data availability

This study used the ProMort datasets, which comprise digitized WSIs of diagnostic core needle biopsies, clinical data from the original diagnosis, and pathologist evaluation and annotations from the re-grading. The image data are available from the corresponding author upon reasonable request. Patient outcome data cannot be shared due to regulatory restrictions.

# Code availability

Core components of the model development and validation pipeline build on open-source software, including PyTorch (https://github.com/pytorch/pytorch) and publicly available implementations of multiple instance learning and vision transformer architectures. The AttentionDeepMIL implementation from AMLab Amsterdam was used for multiple instance learning (https://github.com/AMLab-Amsterdam/AttentionDeepMIL), and the timm library (https://github.com/huggingface/pytorch-image-models) was used for vision transformer backbones. Any preprocessing or analysis scripts used for figure generation are available from the corresponding author upon reasonable request.

# Competing interests

N.M., K.K. and M.E. are shareholders of Clinsight AB. Other authors declare no competing interests. The funder (Karolinska Institutet) did not influence the results/outcomes of the study despite author affiliations with the funder.

# Acknowledgements

K.K. received funding from the SciLifeLab & Wallenberg Data Driven Life Science Program (KAW 2024.0159), Swedish Cancer Society, Instrumentarium Science Foundation and Karolinska Institutet Research Foundation. The construction of the Promort 2 study was supported by grants from the Strategic Research Programme in Epidemiology at Karolinska

Institutet and the Swedish Prostate Cancer Federation, received by A.P. Computations were enabled by the National Academic Infrastructure for Supercomputing in Sweden (NAISS) and the Swedish National Infrastructure for Computing (SNIC) at C3SE partially funded by the Swedish Research Council through grant agreement no. 2022–06725 and no. 2018–05973, and by the supercomputing resource Berzelius provided by the National Supercomputer Centre at Linköping University and the Knut and Alice Wallenberg Foundation.

# Contributions

Study design: X.J., R.Z., Os.A., N.M., P.H.V., M.E., Ol.A., K.K.
Data collection, curation and annotation:  R.Z., Os.A., M.F., F.G., L.M., L.R., A.P.
Data management and software:  X.J., R.Z., N.M., S.E.B., A.P., P.H.V., Ol.A
Drafting of the paper: X.J., R.Z., Os.A., P.H.V., M.E., Ol.A., K.K.

All authors have read and approved the final manuscript. Kimmo Kartasalo is the guarantor.

# Ethics declarations

The study is conducted in agreement with the Declaration of Helsinki and approved by the Swedish Regional Ethics Review Board and the Swedish Ethical Review Authority (permits 2012/1586–31/1, 2016/613–31/2, 2019–01395, 2019–05220). Data were obtained from the National Prostate Cancer Register of Sweden (NPCR), which operates under the Swedish Patient Data Act (Patientdatalagen 2008:355, Chapter 7) and the EU General Data Protection Regulation (GDPR), without reliance on individual informed consent.

# Figures and Tables

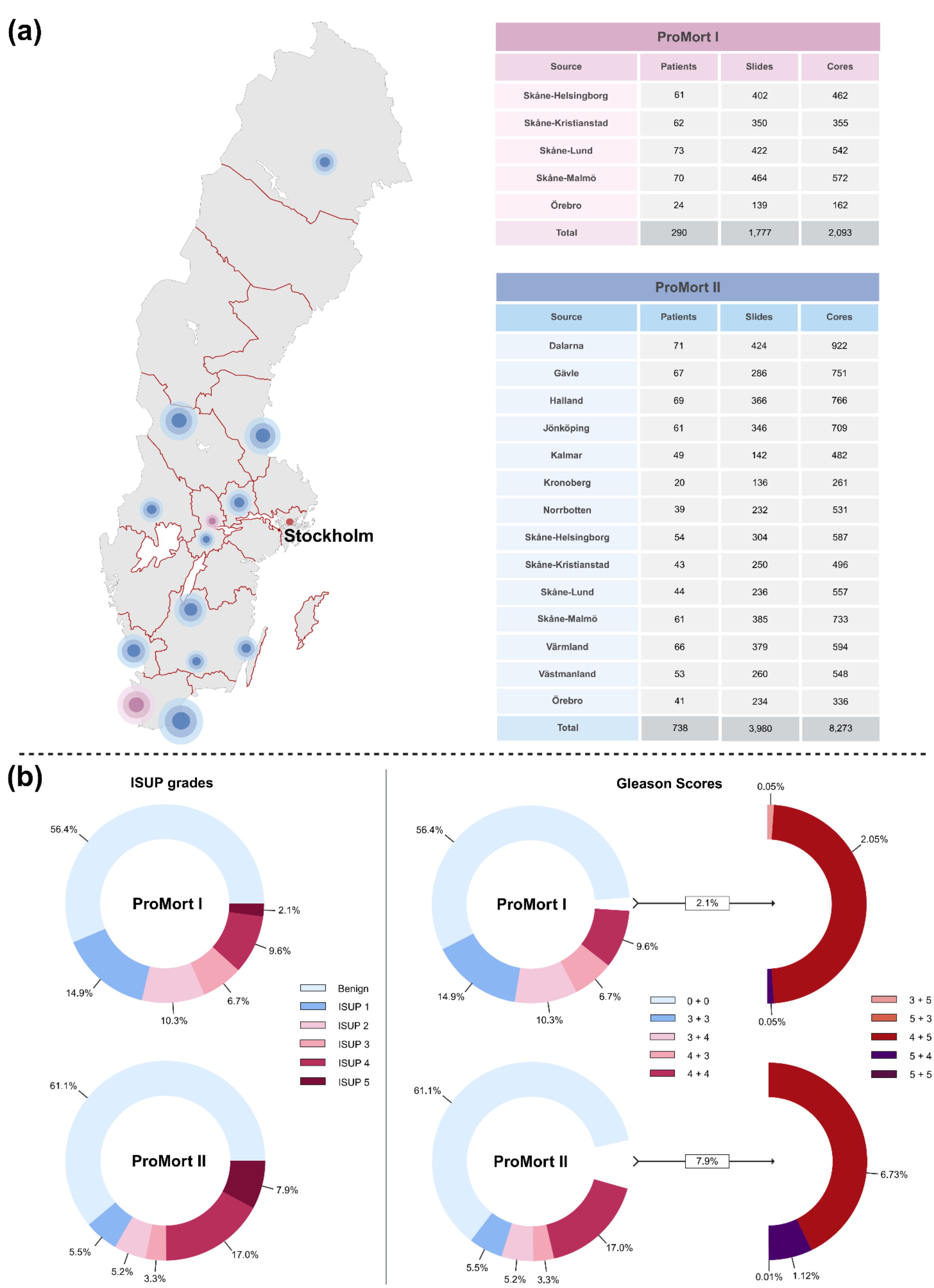

| ProMort I | | | |
|---|---|---|---|
| Source | Patients | Slides | Cores |
| Skåne-Helsingborg | 61 | 402 | 462 |
| Skåne-Kristianstad | 62 | 350 | 355 |
| Skåne-Lund | 73 | 422 | 542 |
| Skåne-Malmö | 70 | 464 | 572 |
| Örebro | 24 | 139 | 162 |
| Total | 290 | 1,777 | 2,093 |

| ProMort II | | | |
|---|---|---|---|
| Source | Patients | Slides | Cores |
| Dalarna | 71 | 424 | 922 |
| Gävle | 67 | 286 | 751 |
| Halland | 69 | 366 | 766 |
| Jönköping | 61 | 346 | 709 |
| Kalmar | 49 | 142 | 482 |
| Kronoberg | 20 | 136 | 261 |
| Norrbotten | 39 | 232 | 531 |
| Skåne-Helsingborg | 54 | 304 | 587 |
| Skåne-Kristianstad | 43 | 250 | 496 |
| Skåne-Lund | 44 | 236 | 557 |
| Skåne-Malmö | 61 | 385 | 733 |
| Värmland | 66 | 379 | 594 |
| Västmanland | 53 | 260 | 548 |
| Örebro | 41 | 234 | 336 |
| Total | 738 | 3,980 | 8,273 |

**Figure 1. Overview of validation dataset with primary reference standard.** In total, 290 patients from ProMort I and 738 patients from ProMort II were included in this validation study after excluding slides not stained with H&E and those with annotation errors. These patients contributed

2,093 cores (from 1,777 slides) in ProMort I and 8,273 cores (from 3,980 slides) in ProMort II that were used as input to the AI model for prostate cancer diagnosis and grading. The numbers presented represent cores annotated by the primary reference pathologist (F.G.) and exclude 3 cores from ProMort II where the segmentation model failed to detect any tissue. **(a)** Geographic distribution of participating sites showing the number of patients, WSIs, and cores per Swedish county (Swedish: *län*). For Skåne county, data are presented at the municipality (Swedish: *kommun*) level due to the large sample size. **(b)** Core-level distribution of ISUP grades and Gleason scores across the ProMort I and ProMort II datasets, where cores with Gleason pattern 5 are shown with expanded detail for better visualization. In ProMort I, Gleason 3+3 and 3+4 accounted for most cancer-positive cores, whereas ProMort II, encompassing the full spectrum of non-metastatic prostate cancer, showed a broader grade distribution with higher grade cores (ISUP 4–5) more frequently represented. The distributions are consistent with the original selection criteria of each dataset. The distribution of cancer grade classes per geographic region is provided in **Extended Table 1**. Abbreviations: H&E=hematoxylin and eosin, ISUP=International Society of Urological Pathology, WSI=whole slide image.

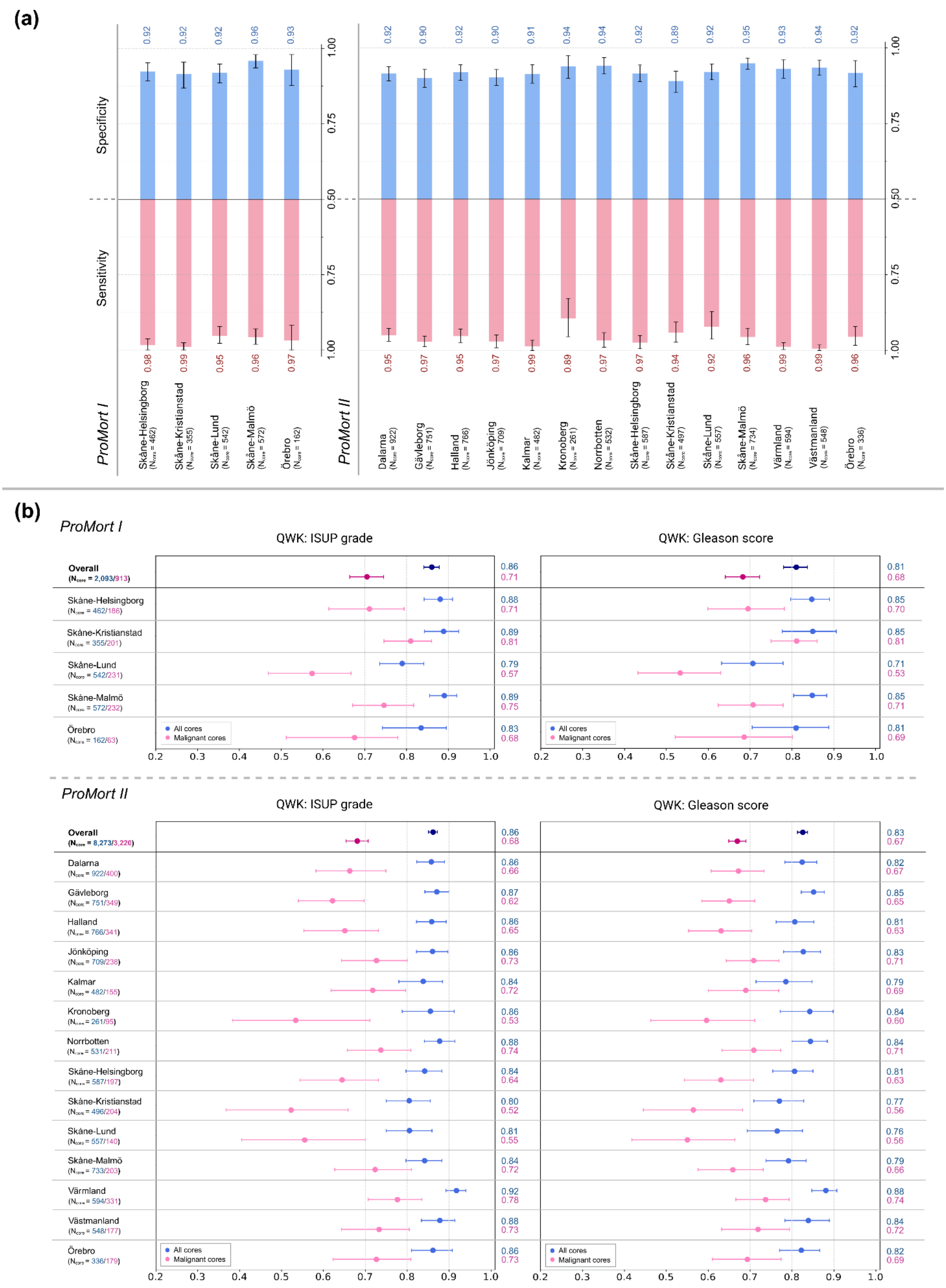

**Figure 2. AI model performance for prostate cancer detection and grading on ProMort I and ProMort II datasets.** Performance metrics are presented evaluating the agreements between AI predictions and the pathologist reference standard (F.G.) across different Swedish regions at the core level. **(a)** Sensitivity and specificity with 95% CI for cancer detection across individual Swedish

counties and municipalities, where $N_{core}$ is the number of cores included in each data source. **(b)** Points indicate QWK values with 95% CI shown as error bars. Blue metrics represent performance on all cores, while pink metrics represent performance restricted to malignant cores as determined by the reference pathologist. $N_{core}$ is the number of cores and malignant cores included in each data source with corresponding color, and the overall dataset performance is provided at the top of each panel as a benchmark. Abbreviation: ISUP=International Society of Urological Pathology, QWK=quadratic weighted kappa, CI=confidence intervals.

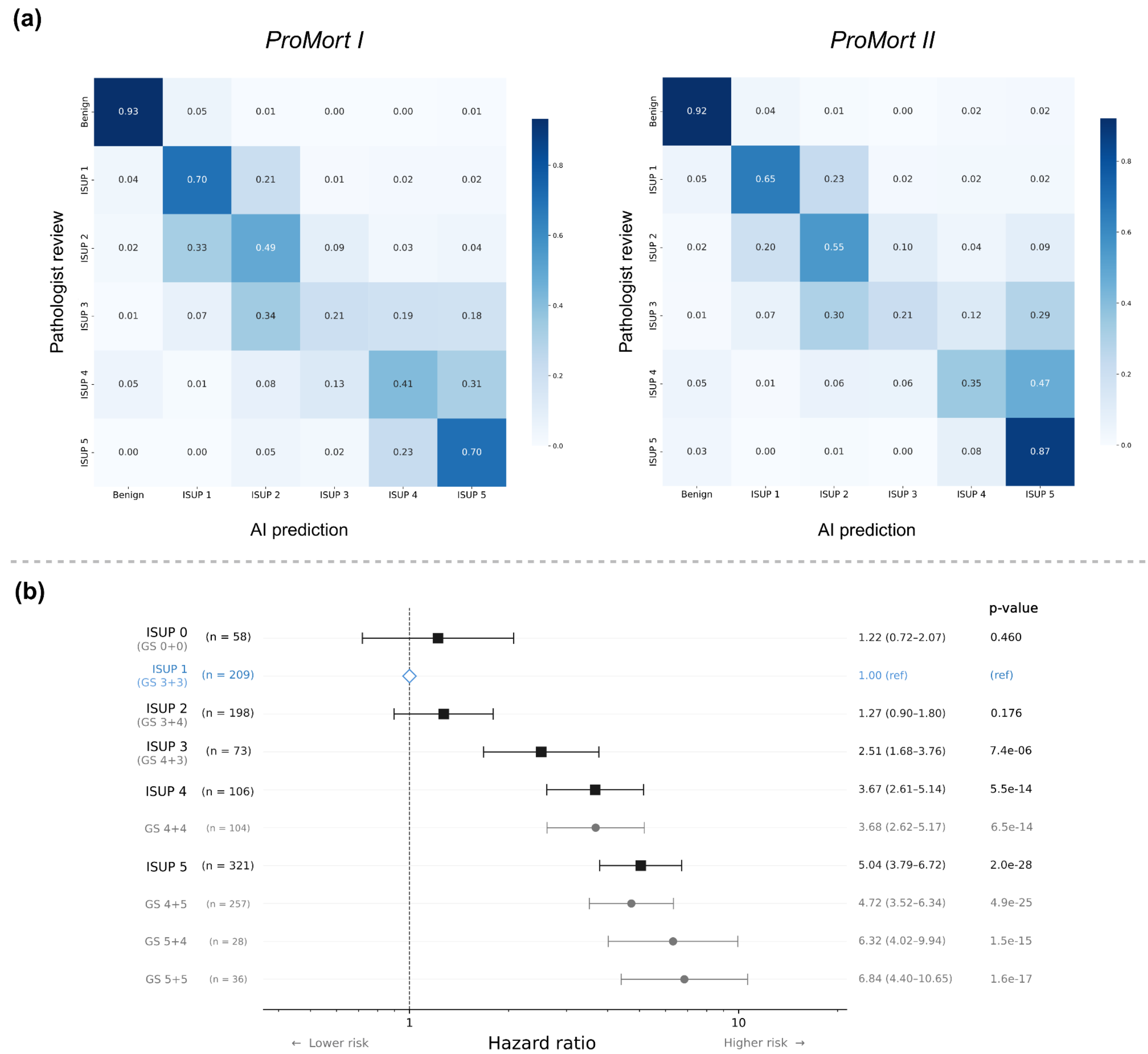

**Figure 3. (a) Confusion matrices showing ISUP grade concordance between AI model predictions and the reference pathologist annotations.** Values are row-normalized to show the proportion of pathologist-graded cores assigned to each AI-predicted category. Darker blue indicates higher concordance between AI and pathologist assessments. **(b) Hazard ratios for prostate cancer–specific mortality by AI-predicted patient-level grade.** Values are estimated using Cox proportional hazards regression with ISUP 1 (GS 3+3) as reference. Point estimates with 95% CI and *p*-values are shown for each grade group. For ISUP 4 and 5, Gleason score subgroups are shown separately. GS 3+5 (n=2) was excluded from the subgroup display due to insufficient sample size. n denotes the number of patients per group. Abbreviations: CI=confidence interval, GS=Gleason score, ISUP=International Society of Urological Pathology.

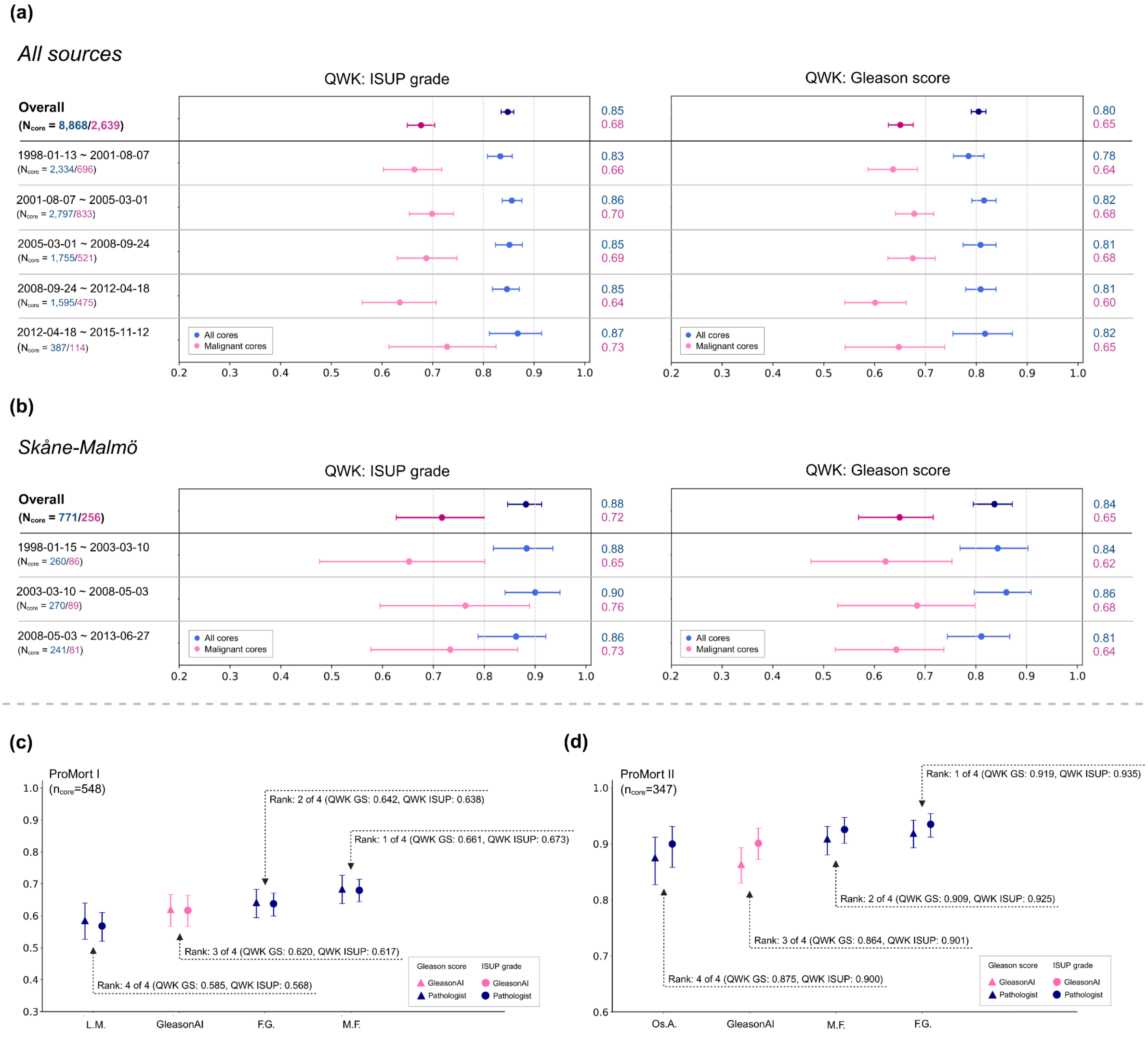

**Figure 4. Sensitivity analyses for AI model performance. (a,b) AI performance for prostate cancer grading across different sample collection periods.** The temporal robustness of the AI model performance is evaluated with stratified sampling to achieve comparable ISUP grade distributions across time bins. ProMort I and II datasets were merged for analysis. Panels **(a)** show performance across all geographic sources; panels **(b)** show performance for Skåne-Malmö only. Left column shows ISUP grade concordance, right column shows Gleason score concordance. Blue metrics represent all cores, pink metrics represent malignant cores only (as determined by reference pathologist F.G.). **(c-d) Inter-observer variability in prostate cancer grading.** Comparison of grading concordance between individual observers (pathologists and *GleasonAI*) versus all other observers on **(c)**: 548 cores from ProMort I annotated by F.G., L.M., and M.F., and on **(d)**: 347 cores from ProMort II annotated by F.G., M.F., and Os.A. *GleasonAI* points show mean concordance between AI and all pathologists; pathologist points show mean concordance with other pathologists. All points show QWK values with 95% CI. Observers ranked by ISUP grade QWK performance. Abbreviations: CI=confidence interval, ISUP=International Society of Urological Pathology, QWK=quadratic weighted kappa.

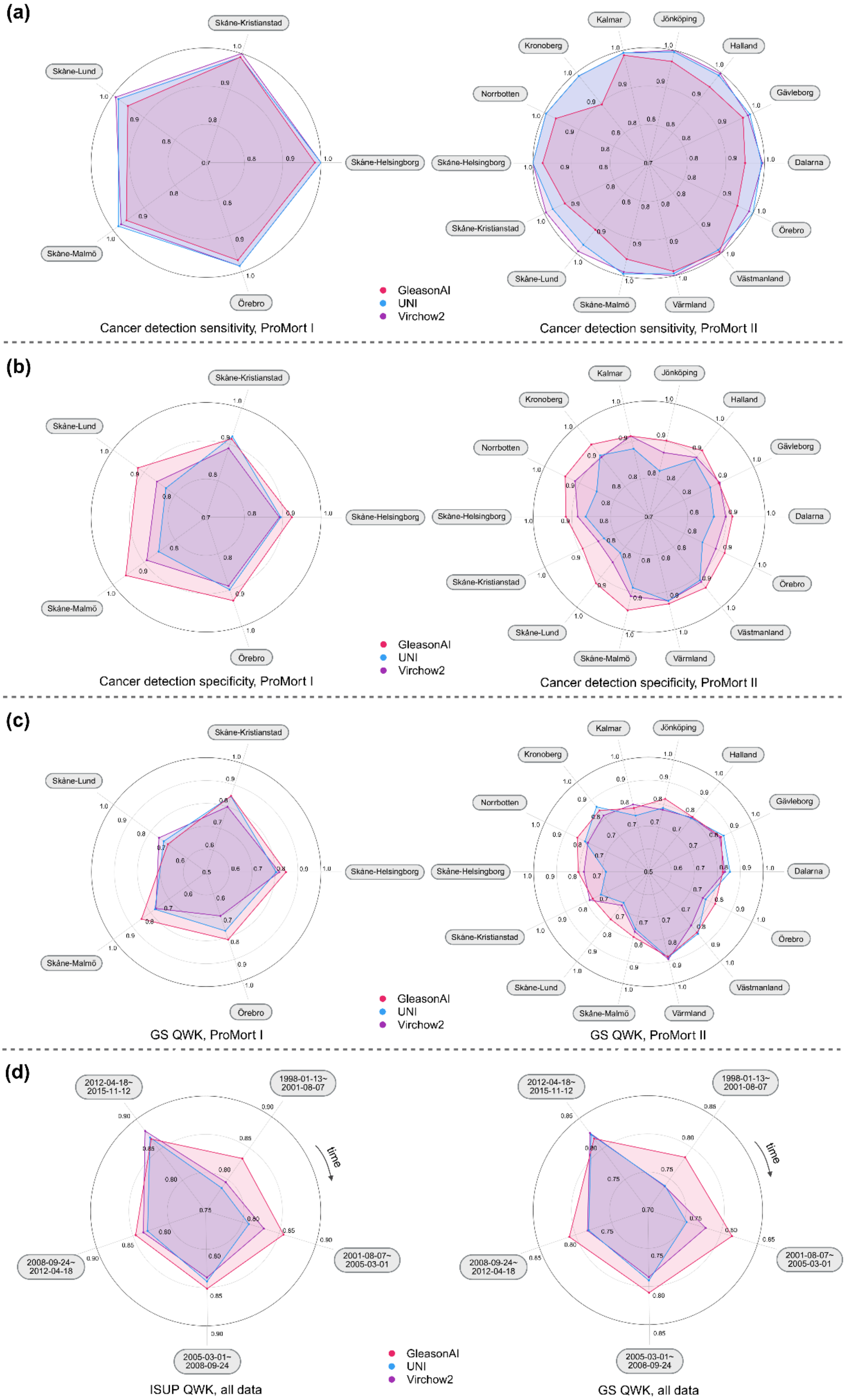
(a)
Skåne-Kristianstad
Skåne-Lund
Skåne-Helsingborg
Skåne-Malmö
Örebro
GleasonAI
UNI
Virchow2
Cancer detection sensitivity, ProMort I
Kalmar
Jönköping
Kronoberg
Halland
Norrbotten
Gävleborg
Dalarna
Örebro
Västmanland
Värmland
Skåne-Malmö
Skåne-Lund
Skåne-Kristianstad
Skåne-Helsingborg
Cancer detection sensitivity, ProMort II
(b)
Cancer detection specificity, ProMort I
Cancer detection specificity, ProMort II
(c)
GS QWK, ProMort I
GS QWK, ProMort II
(d)
2012-04-18~ 2015-11-12
1998-01-13~ 2001-08-07
2001-08-07~ 2005-03-01
2005-03-01~ 2008-09-24
2008-09-24~ 2012-04-18
time
ISUP QWK, all data
GS QWK, all data

**Figure 5. Performance comparison of *GleasonAI* against foundation models.** Radar plots are presented comparing the end-to-end ABMIL model *GleasonAI* with general-purpose pathology foundation models *UNI* and *Virchow2* using identical training and test datasets. For panels **(a-c)**, the left column shows ProMort I results and the right column shows ProMort II results, and each axis represents a geographic region with performance values. **(a)** Cancer detection sensitivity. **(b)** Cancer detection specificity. **(c)** Gleason score concordance measured by QWK for all cores. **(d)** Temporal comparison of grading concordance across five stratified time intervals, showing ISUP grade and Gleason score QWK for all cores from ProMort I and II. Axes are arranged clockwise from the oldest (1998–2001) to the most recent (2012-2015) biopsy collection period. Abbreviation: ABMIL=Attention-based multiple instance learning, GS=Gleason score, ISUP=International Society of Urological Pathology, QWK=quadratic weighted kappa.

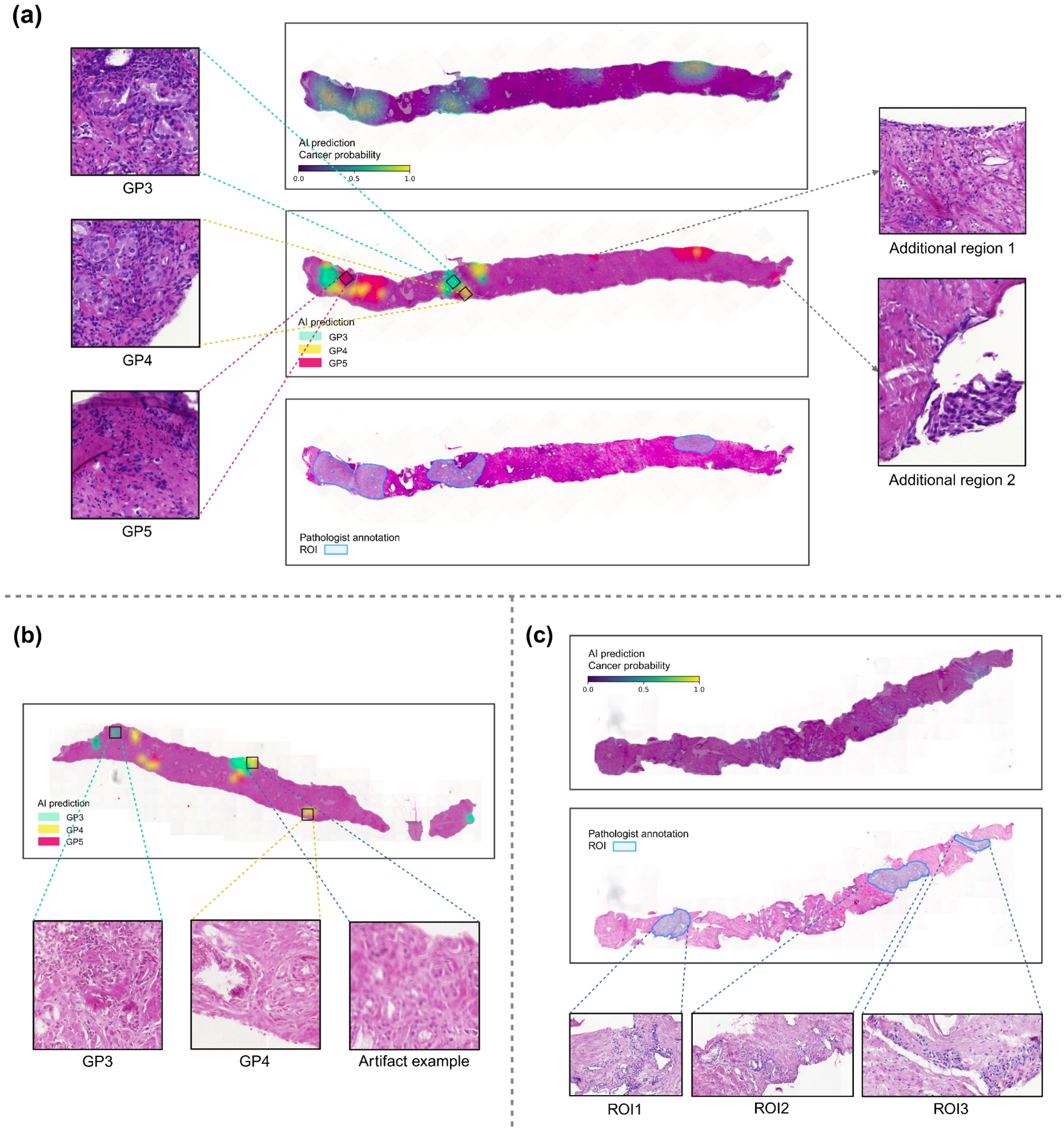

**Figure 6. Visualization of AI model predictions compared with pathologist annotations in prostate core biopsies.** Three representative core biopsies demonstrating the patch-level prediction capabilities of the weakly-supervised AI model trained on WSI level Gleason scores. **(a)** Core showing agreement between AI predictions and pathologist assessment. Top: heatmap of AI-predicted cancer probability per patch (256×256 pixels). Middle: heatmap of AI-predicted GPs, showing the most probable pattern per patch. Bottom: reference pathologist annotation with regions of interest (ROI) marked in light blue indicating cancerous areas. Left column shows representative patches of Gleason patterns 3–5 as predicted by the AI model and confirmed by reviewing pathologist Os.A. Right column shows: Additional region 1 contains small cancer foci initially missed by the reference pathologist; Additional region 2 shows possible tissue contamination from processing that the AI correctly identified as cancerous tissue despite being clinically disregarded. **(b)** Core demonstrating AI robustness to image quality artifacts, where the AI assigned Gleason score 3+4 while the reference pathologist assigned 0+0. The magnified regions are representative patches of GP3 and GP4 confirmed by Os.A., alongside an example of blur artifacts present throughout this core that likely

contributed to the initial benign assessment. **(c)** Core showing AI-pathologist disagreement where the reference pathologist assigned 4+3 and the AI predicted 0+0. Top: AI-predicted cancer probability heatmap showing elevated probability in the region corresponding to ROI3, though below the classification threshold. Bottom: pathologist-annotated cancerous regions with three ROIs shown in detail below; reviewing pathologist Os.A. confirmed ROI1 and ROI2 as benign, while ROI3 represents a borderline case that would require immunohistochemistry for definitive diagnosis in clinical practice. Notably, the AI model assigned appropriate uncertainty quantification for this ambiguous region. These examples demonstrate that weakly-supervised training enables the AI model to learn patch-level Gleason pattern recognition and maintain robust performance despite image quality variations, while also highlighting expected variability in borderline cases that mirrors inter-observer variation in clinical practice. Abbreviation: GP=Gleason pattern, ROI=region of interest, WSI=whole slide image

## Supplementary Appendix

# Validation of an AI-based end-to-end model for prostate pathology using long-term archived routine samples

Xiaoyi Ji[1], Renata Zelic[2,3], Oskar Aspegren[2,3,4], Nita Mulliqi[5], Michelangelo Fiorentino[6], Francesca Giunchi[7], Luca Molinaro[8], Sol Erika Boman[1,2], Lorenzo Richiardi[9,10], Andreas Pettersson[3,11], Per Henrik Vincent[2,3], Martin Eklund[1], Olof Akre[2,3], Kimmo Kartasalo[5]

1. Department of Medical Epidemiology and Biostatistics, Karolinska Institutet, Stockholm, Sweden
2. Department of Molecular Medicine and Surgery, Karolinska Institutet, Stockholm, Sweden
3. Department of Pelvic Cancer, Cancer Theme, Karolinska University Hospital, Stockholm, Sweden
4. Department of Pathology and Cancer Diagnostics, Karolinska University Hospital, Stockholm, Sweden
5. Department of Medical Epidemiology and Biostatistics, SciLifeLab, Karolinska Institutet, Stockholm, Sweden
6. Department of Medical and Surgical Sciences, University of Bologna, Bologna, Italy
7. Department of Pathology, IRCCS Azienda Ospedaliero-Universitaria di Bologna, Bologna, Italy
8. Division of Pathology, AOU Città Della Salute e Della Scienza di Torino, Turin, Italy
9. Department of Medical Sciences, University of Turin, Torino, Italy
10. Cancer Epidemiology Unit, University Hospital Città della Scienza e della Salute di Torino and CPO-Piemonte, Torino, Italy
11. Clinical Epidemiology Division, Department of Medicine Solna, Karolinska Institutet, Stockholm, Sweden.

# Extended Figures

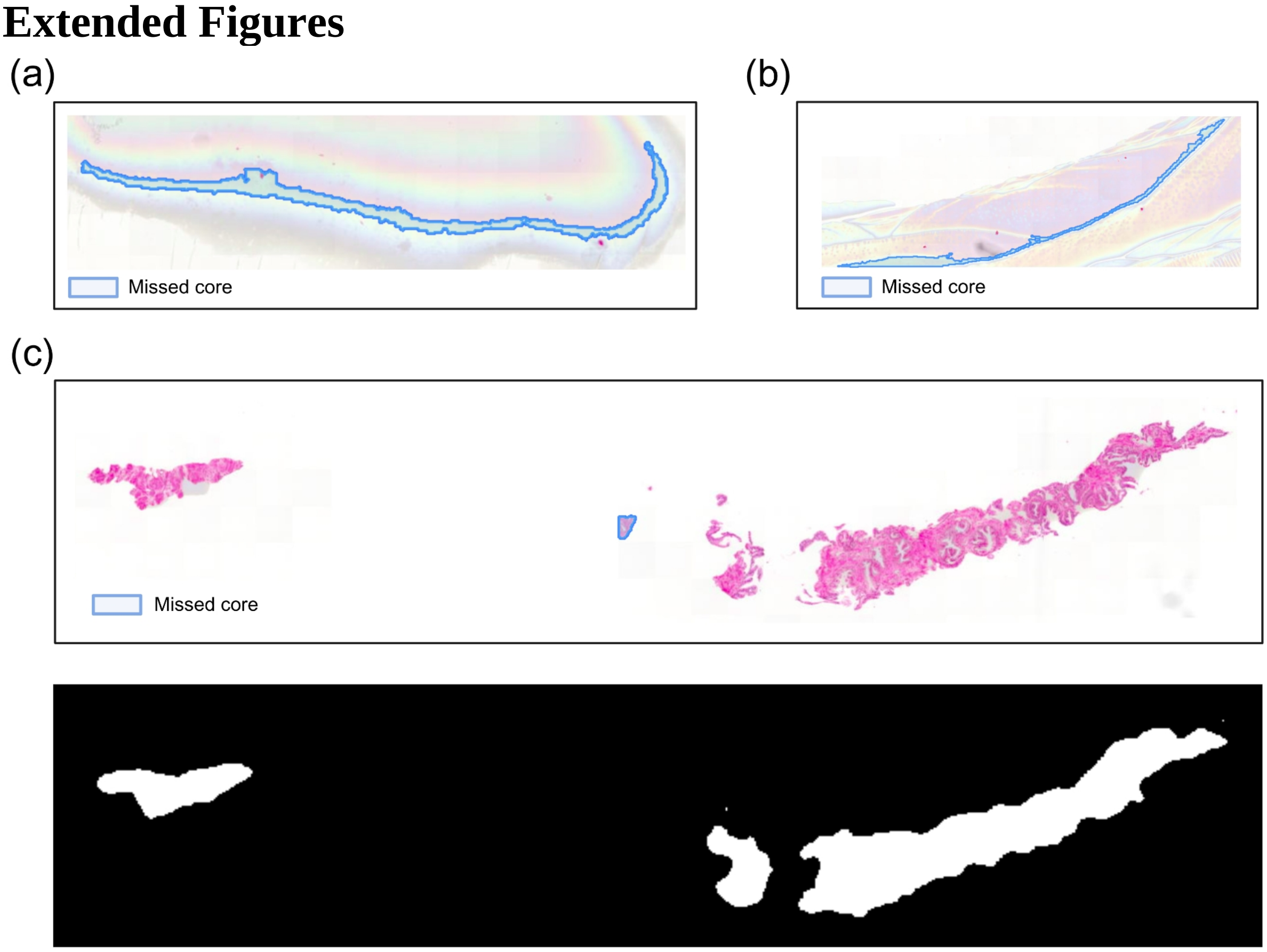

**Extended Figure 1. All three pre-processing failures in the AI-based tissue segmentation process. (a, b)** Annotation errors where regions marked by the pathologist (blue outlines) do not represent actual tissue cores. **(c)** Tissue fragment missed by the segmentation model: the pathology image (upper panel) shows a portion of a WSI with the blue outline indicating a tissue fragment that was not detected by the AI model. The corresponding segmentation mask (lower panel) displays the model's output, where white regions represent areas identified as tissue and black regions represent background. The absence of white pixels in the location corresponding to the blue-outlined region confirms the segmentation failure. While labeled as a missed "core" for consistency with core-level analysis, this represents a small tissue fragment rather than a complete core (see study protocol for detailed explanation).

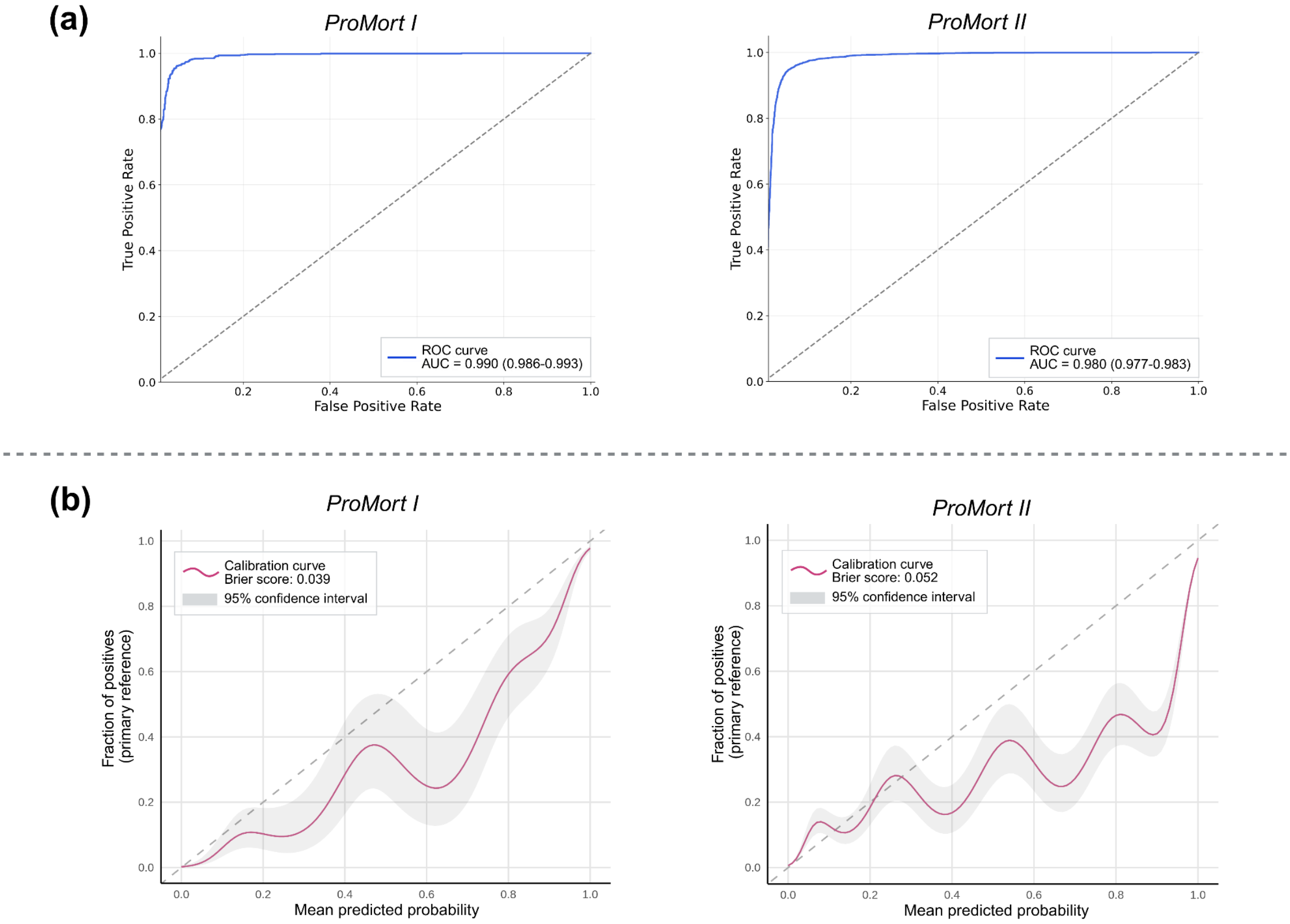

**Extended Figure 2. Cancer diagnosis evaluation for *GleasonAI* on ProMort I** ($n_{core}$=2,093) **and ProMort II** ($n_{core}$=8,273)**. (a) ROC curve.** The diagonal dashed line represents prediction by chance. The AUC is 0.990 (0.986-0.993) and 0.980 (0.977-0.983) for ProMort I and II, respectively. **(b) Calibration plots assessing the alignment between the predicted cancer probabilities from the model and the observed cancer frequencies across probability ranges.** The diagonal dashed line represents perfect calibration, where predicted probabilities match observed frequencies. The calibration curve was fitted using a generalized additive model (GAM) with a smooth term (via the *cal_plot_logistic* function in the R *probably* package), with shaded areas indicating 95% confidence intervals. Brier scores, which reflect overall predictive performance combining discrimination and calibration, were 0.039 and 0.052 for ProMort I and II, respectively.

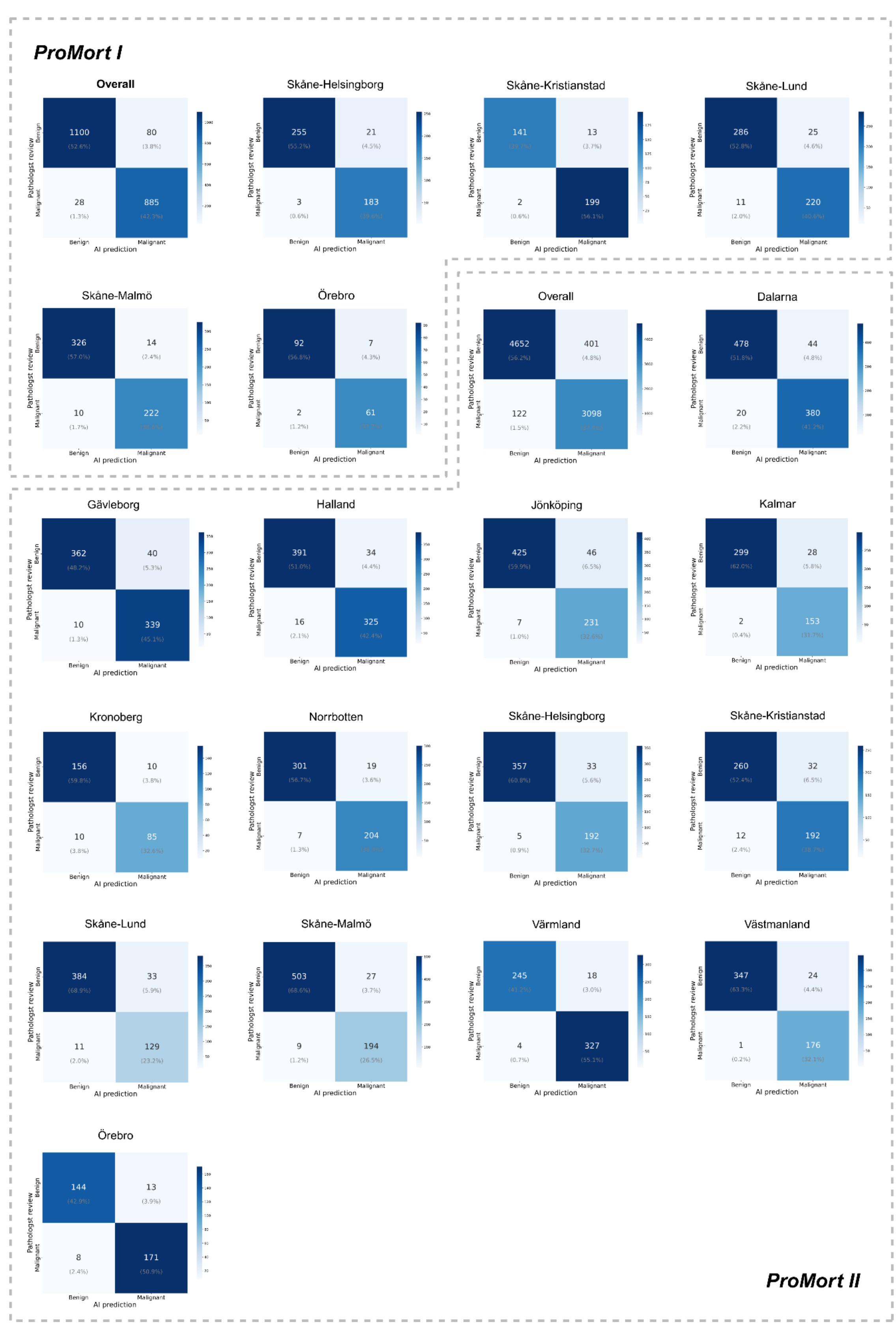

**Extended Figure 3. Confusion matrices for cancer detection for each data source.** Values are numbers and percentages of pathologist-diagnosed cores assigned to benign and malignant categories. Darker blue indicates higher concordance between AI and pathologist assessments.

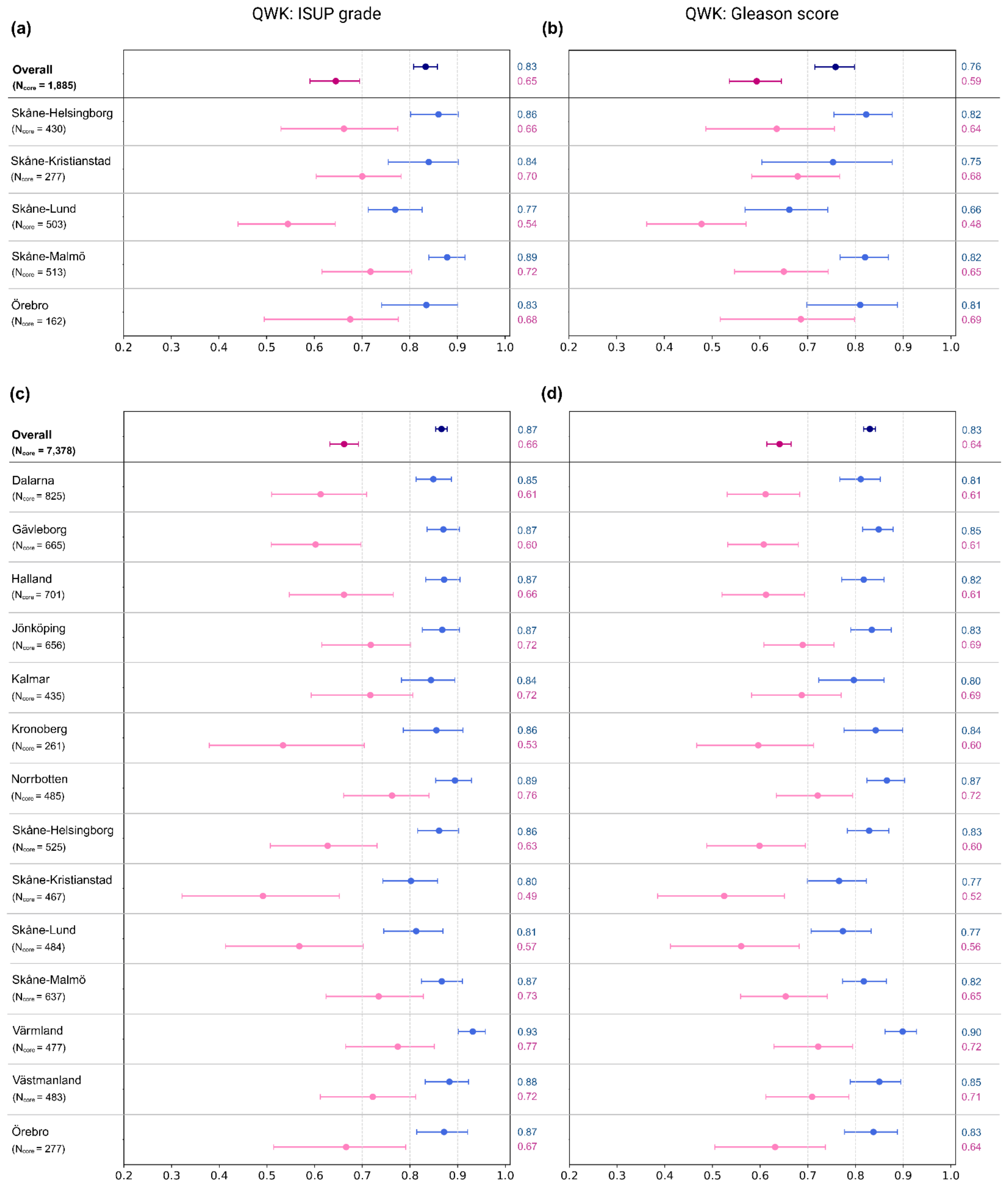

**Extended Figure 4. AI model performance for Gleason grading across different geographic regions.** Sensitivity analysis is presented evaluating model performance robustness over data sources with stratified sampling to achieve comparable ISUP grade distributions across different Swedish counties and municipalities, to confirm that these regional performance patterns reflected true model generalization rather than confounding due to differences in grade distribution across sites. The left column (a, c) shows ISUP grade concordance and the right column (b, d) shows Gleason score concordance. **(a, b)** ProMort I dataset. **(c, d)** ProMort II dataset. Blue metrics represent performance on all cores, while pink metrics represent performance restricted to malignant cores as determined by the reference pathologist (F.G.). Points indicate QWK values with 95% confidence intervals shown as error bars. The $N_{core}$ indicates the number of cores included in each data source, and the overall dataset performance is provided at the top of each panel as a benchmark. Abbreviation: ISUP=International Society of Urological Pathology, QWK=quadratic weighted kappa.

# Extended Tables

| Source | Core no. | Cancer grade classes | | | | | | | | |
|---|---|---|---|---|---|---|---|---|---|---|
| | | ISUP 0 | ISUP 1 | ISUP 2 | ISUP 3 | ISUP 4 | | ISUP 5 | | |
| | | 0+0 | 3+3 | 3+4 | 4+3 | 4+4 | 3+5 | 4+5 | 5+4 | 5+5 |
| **ProMort I** | | | | | | | | | | |
| Skåne-Helsingborg | 462 (186) | 53.5% | 20.3% | 16.2% | 2.8% | 4.3% | 0 | 2.6% | 0.2% | 0 |
| Skåne-Kristianstad | 355 (201) | 40.6% | 18.0% | 12.7% | 10.1% | 10.1% | 0.3% | 7.0% | 0.3% | 0.8% |
| Skåne-Lund | 542 (231) | 48.3% | 18.1% | 12.2% | 5.2% | 7.2% | 0.2% | 7.6% | 0.2% | 1.1% |
| Skåne-Malmö | 572 (232) | 51.4% | 18.2% | 9.6% | 5.0% | 5.8% | 0.5% | 8.0% | 1.0% | 0.5% |
| Örebro | 162 (63) | 55.6% | 21.0% | 16.7% | 2.5% | 1.9% | 0.6% | 1.9% | 0 | 0 |
| **Overall** | 2,093 (913) | 49.5% | 18.8% | 12.8% | 5.2% | 6.3% | 0.3% | 6.1% | 0.4% | 0.6% |
| **ProMort II** | | | | | | | | | | |
| Dalarna | 922 (400) | 49.5% | 9.8% | 10.2% | 1.6% | 8.2% | 0.1% | 13.8% | 2.7% | 4.1% |
| Gävleborg | 751 (349) | 47.3% | 7.3% | 9.3% | 3.1% | 10.9% | 0.1% | 17.3% | 2.5% | 2.1% |
| Halland | 766 (341) | 49.7% | 7.8% | 6.9% | 2.2% | 5.2% | 5.2% | 17.0% | 6.3% | 4.3% |
| Jönköping | 709 (238) | 54.7% | 11.0% | 6.3% | 5.8% | 6.9% | 0.3% | 9.9% | 2.5% | 2.5% |
| Kalmar | 482 (155) | 60.0% | 6.6% | 9.5% | 1.9% | 5.0% | 1.0% | 11.6% | 1.9% | 2.5% |
| Kronoberg | 261 (95) | 57.9% | 4.2% | 6.5% | 1.9% | 14.9% | 0.4% | 8.8% | 1.5% | 3.8% |
| Norrbotten | 531 (211) | 51.4% | 12.4% | 5.6% | 1.1% | 5.1% | 1.5% | 18.1% | 2.1% | 2.6% |
| Skåne-Helsingborg | 587 (197) | 57.4% | 11.2% | 11.1% | 2.7% | 7.7% | 0 | 5.3% | 2.2% | 2.4% |
| Skåne-Kristianstad | 496 (204) | 49.8% | 4.8% | 6.5% | 2.8% | 18.1% | 0.6% | 13.1% | 1.6% | 2.6% |
| Skåne-Lund | 557 (140) | 62.1% | 7.0% | 6.5% | 0.9% | 6.5% | 0 | 12.4% | 1.8% | 2.9% |
| Skåne-Malmö | 733 (203) | 64.4% | 9.0% | 3.7% | 1.8% | 5.9% | 1.1% | 10.2% | 1.1% | 2.9% |
| Värmland | 594 (331) | 41.2% | 9.8% | 8.6% | 5.1% | 12.3% | 0.5% | 18.2% | 3.5% | 0.8% |
| Västmanland | 548 (177) | 62.0% | 10.8% | 8.6% | 4.4% | 4.9% | 0.2% | 7.3% | 1.1% | 0.7% |
| Örebro | 336 (179) | 40.0% | 11.3% | 9.5% | 1.2% | 7.4% | 0.6% | 20.8% | 5.1% | 4.2% |
| **Overall** | 8,273 (3,220) | 53.4% | 9.0% | 7.8% | 2.7% | 8.2% | 0.5% | 13.2% | 2.6% | 2.8% |

**Extended Table 1. Core-level distribution of cancer grades across geographic regions in ProMort I and ProMort II datasets.** Numbers under the ISUP are the Gleason scores associated with the ISUP grades. Core numbers are shown with the number of malignant cores in parentheses. Abbreviation: ISUP=International Society of Urological Pathology.

| Source | Core no. | ISUP LWK | Malignant ISUP LWK | ISUP QWK | Malignant ISUP QWK | ISUP C-index | Malignant ISUP C-index |
|---|---|---|---|---|---|---|---|
| **ProMort I** | | | | | | | |
| Skåne-Helsingborg | 462 (186) | 0.774 (0.735-0.811) | 0.556 (0.480-0.628) | 0.880 (0.840-0.912) | 0.711 (0.619-0.788) | 0.943 (0.927-0.957) | 0.826 (0.788-0.859) |
| Skåne-Kristianstad | 355 (201) | 0.797 (0.756-0.832) | 0.665 (0.591-0.726) | 0.889 (0.742-0.923) | 0.810 (0.747-0.860) | 0.938 (0.920-0.955) | 0.870 (0.838-0.897) |
| Skåne-Lund | 542 (231) | 0.706 (0.662-0.746) | 0.471 (0.398-0.545) | 0.790 (0.733-0.836) | 0.574 (0.477-0.665) | 0.910 (0.889-0.930) | 0.760 (0.711-0.806) |
| Skåne-Malmö | 572 (232) | 0.807 (0.774-0.838) | 0.628 (0.561-0.694) | 0.890 (0.854- 0.921) | 0.746 (0.660-0.820) | 0.944 (0.927-0.960) | 0.850 (0.806-0.885) |
| Örebro | 162 (63) | 0.725 (0.652-0.793) | 0.461 (0.305-0.592) | 0.835 (0.737- 0.900) | 0.675 (0.519-0.783) | 0.924 (0.889-0.956) | 0.780 (0.693-0.845) |
| **Overall** | 2,093 (913) | 0.770 (0.750-0.788) | 0.574 (0.538-0.610) | 0.860 (0.838-0.881) | 0.705 (0.663-0.745) | 0.932 (0.923-0.940) | 0.815 (0.796-0.833) |
| **ProMort II** | | | | | | | |
| Dalarna | 922 (400) | 0.797 (0.765-0.828) | 0.586 (0.530-0.644) | 0.858 (0.823-0.889) | 0.663 (0.585-0.734) | 0.917 (0.900-0.932) | 0.813 (0.783-0.842) |
| Gävleborg | 751 (349) | 0.785 (0.752-0.813) | 0.502 (0.435-0.559) | 0.871 (0.838-0.898) | 0.621 (0.538-0.693) | 0.922 (0.906-0.937) | 0.796 (0.767-0.825) |
| Halland | 766 (341) | 0.786 (0.755-0.818) | 0.543 (0.483-0.608) | 0.859 (0.820-0.893) | 0.651 (0.559-0.742) | 0.911 (0.892-0.928) | 0.779 (0.739-0.818) |
| Jönköping | 709 (238) | 0.780 (0.744-0.814) | 0.576 (0.506-0.638) | 0.861 (0.821-0.898) | 0.726 (0.645-0.794) | 0.937 (0.921-0.951) | 0.834 (0.804-0.863) |
| Kalmar | 482 (155) | 0.768 (0.722-0.812) | 0.567 (0.475-0.642) | 0.838 (0.782-0.886) | 0.718 (0.618-0.795) | 0.939 (0.921-0.955) | 0.820 (0.775-0.859) |
| Kronoberg | 261 (95) | 0.792 (0.724-0.848) | 0.489 (0.360-0.601) | 0.855 (0.789-0.909) | 0.534 (0.373-0.681) | 0.913 (0.877-0.944) | 0.815 (0.761-0.864) |
| Norrbotten | 531 (211) | 0.801 (0.763-0.837) | 0.594 (0.521-0.661) | 0.877 (0.838-0.912) | 0.737 (0.653-0.807) | 0.929 (0.913-0.945) | 0.801 (0.766-0.835) |
| Skåne-Helsingborg | 587 (197) | 0.765 (0.721-0.804) | 0.540 (0.461-0.617) | 0.841 (0.791-0.883) | 0.645 (0.554-0.732) | 0.936 (0.919-0.953) | 0.805 (0.756-0.849) |
| Skåne-Kristianstad | 496 (204) | 0.758 (0.709-0.803) | 0.508 (0.408-0.595) | 0.805 (0.748-0.853) | 0.523 (0.375-0.665) | 0.896 (0.866-0.920) | 0.776 (0.718-0.828) |
| Skåne-Lund | 557 (140) | 0.738 (0.688-0.788) | 0.472 (0.368-0.598) | 0.805 (0.746-0.859) | 0.555 (0.409-0.760) | 0.918 (0.890-0.940) | 0.770 (0.713-0.825) |
| Skåne-Malmö | 733 (203) | 0.784 (0.741-0.823) | 0.605 (0.531-0.672) | 0.842 (0.794-0.885) | 0.723 (0.623-0.801) | 0.935 (0.915-0.951) | 0.811 (0.766-0.851) |
| Värmland | 594 (331) | 0.841 (0.810-0.868) | 0.656 (0.595-0.708) | 0.918 (0.889-0.941) | 0.777 (0.708-0.834) | 0.934 (0.918-0.948) | 0.834 (0.798-0.864) |
| Västmanland | 548 (177) | 0.808 (0.767-0.844) | 0.612 (0.529-0.688) | 0.877 (0.834-0.916) | 0.733 (0.647-0.806) | 0.953 (0.939-0.966) | 0.823 (0.778-0.865) |
| Örebro | 336 (179) | 0.785 (0.740-0.830) | 0.609 (0.525-0.684) | 0.862 (0.811-0.908) | 0.727 (0.627-0.809) | 0.905 (0.881-0.929) | 0.808 (0.762-0.851) |
| **Overall** | 8,273 (3,220) | 0.792 (0.782-0.802) | 0.573 (0.553-0.593) | 0.862 (0.850-0.872) | 0.681 (0.655-0.706) | 0.926 (0.921-0.931) | 0.805 (0.796-0.816) |

**Extended Table 2. AI model performance for ISUP grade prediction across geographic regions in ProMort I and ProMort II datasets.** Subgroup analysis is presented showing agreements between AI predictions and the pathologist reference standard across different Swedish counties and municipalities by LWK, QWK, and with 95% confidence intervals. Metrics are presented for all cores and for malignant cores only (cores diagnosed as malignant by the reference standard). Core numbers are shown with the number of malignant cores in parentheses. Abbreviation: C-index=concordance index, ISUP=International Society of Urological Pathology, LWK=linear weighted kappa, QWK=quadratic weighted kappa.

| Source | Core no. | GS LWK | Malignant GS LWK | GS QWK | Malignant GS QWK | GS C-index | Malignant GS C-index |
|---|---|---|---|---|---|---|---|
| **ProMort I** | | | | | | | |
| Skåne-Helsingborg | 462 (186) | 0.745 (0.703-0.788) | 0.545 (0.464-0.623) | 0.848 (0.802-0.889) | 0.695 (0.590-0.784) | 0.943 (0.927-0.957) | 0.826 (0.784-0.858) |
| Skåne-Kristianstad | 355 (201) | 0.768 (0.716-0.810) | 0.658 (0.593-0.714) | 0.850 (0.776-0.907) | 0.811 (0.753-0.855) | 0.938 (0.917-0.954) | 0.871 (0.835-0.902) |
| Skåne-Lund | 542 (231) | 0.652 (0.600-0.703) | 0.442 (0.366-0.513) | 0.707 (0.639-0.777) | 0.533 (0.433-0.624) | 0.909 (0.884-0.931) | 0.761 (0.711-0.805) |
| Skåne-Malmö | 572 (232) | 0.761 (0.721-0.796) | 0.585 (0.520-0.651) | 0.849 (0.804- 0.885) | 0.798 (0.624-0.781) | 0.945 (0.927-0.960) | 0.853 (0.812-0.886) |
| Örebro | 162 (63) | 0.705 (0.622-0.778) | 0.469 (0.313-0.597) | 0.810 (0.697- 0.891) | 0.686 (0.522-0.800) | 0.924 (0.891-0.953) | 0.781 (0.702-0.846) |
| **Overall** | 2,093 (913) | 0.731 (0.710-0.751) | 0.554 (0.517-0.589) | 0.811 (0.778-0.838) | 0.683 (0.638-0.724) | 0.932 (0.923-0.940) | 0.817 (0.797-0.836) |
| **ProMort II** | | | | | | | |
| Dalarna | 922 (400) | 0.746 (0.713-0.774) | 0.546 (0.495-0.596) | 0.824 (0.786-0.858) | 0.672 (0.614-0.732) | 0.918 (0.901-0.934) | 0.823 (0.792-0.852) |
| Gävleborg | 751 (349) | 0.739 (0.709-0.767) | 0.483 (0.426-0.533) | 0.851 (0.824-0.877) | 0.650 (0.583-0.705) | 0.926 (0.912-0.940) | 0.816 (0.789-0.845) |
| Halland | 766 (341) | 0.720 (0.684-0.751) | 0.490 (0.431-0.548) | 0.807 (0.762-0.846) | 0.631 (0.547-0.706) | 0.917 (0.897-0.936) | 0.825 (0.783-0.860) |
| Jönköping | 709 (238) | 0.732 (0.696-0.768) | 0.531 (0.469-0.588) | 0.827 (0.780-0.869) | 0.709 (0.637-0.766) | 0.940 (0.924-0.955) | 0.859 (0.828-0.889) |
| Kalmar | 482 (155) | 0.716 (0.663-0.766) | 0.532 (0.452-0.612) | 0.785 (0.712-0.851) | 0.690 (0.605-0.767) | 0.941 (0.920-0.957) | 0.848 (0.802-0.887) |
| Kronoberg | 261 (95) | 0.752 (0.690-0.810) | 0.469 (0.351-0.573) | 0.842 (0.773-0.901) | 0.596 (0.460-0.717) | 0.915 (0.876-0.949) | 0.834 (0.779-0.885) |
| Norrbotten | 531 (211) | 0.750 (0.714-0.786) | 0.550 (0.481-0.613) | 0.844 (0.805-0.882) | 0.709 (0.634-0.772) | 0.935 (0.918-0.951) | 0.835 (0.801-0.865) |
| Skåne-Helsingborg | 587 (197) | 0.720 (0.676-0.758) | 0.506 (0.431-0.582) | 0.806 (0.754-0.849) | 0.630 (0.542-0.705) | 0.938 (0.919-0.955) | 0.816 (0.771-0.860) |
| Skåne-Kristianstad | 496 (204) | 0.716 (0.663-0.761) | 0.495 (0.405-0.576) | 0.770 (0.702-0.827) | 0.565 (0.438-0.679) | 0.896 (0.870-0.921) | 0.786 (0.728-0.836) |
| Skåne-Lund | 557 (140) | 0.691 (0.636-0.740) | 0.439 (0.349-0.528) | 0.765 (0.696-0.826) | 0.550 (0.431-0.666) | 0.917 (0.890-0.940) | 0.778 (0.714-0.832) |
| Skåne-Malmö | 733 (203) | 0.720 (0.680-0.759) | 0.525 (0.452-0.587) | 0.792 (0.741-0.839) | 0.659 (0.569-0.729) | 0.937 (0.919-0.955) | 0.827 (0.786-0.869) |
| Värmland | 594 (331) | 0.788 (0.757-0.816) | 0.601 (0.539-0.656) | 0.881 (0.848-0.909) | 0.737 (0.672-0.789) | 0.939 (0.923-0.953) | 0.851 (0.817-0.882) |
| Västmanland | 548 (177) | 0.760 (0.717-0.801) | 0.574 (0.498-0.643) | 0.839 (0.787-0.889) | 0.719 (0.636-0.789) | 0.955 (0.941-0.968) | 0.840 (0.794-0.879) |
| Örebro | 336 (179) | 0.726 (0.678-0.769) | 0.549 (0.467-0.622) | 0.822 (0.768-0.869) | 0.693 (0.608-0.764) | 0.914 (0.886-0.936) | 0.837 (0.789-0.875) |
| **Overall** | 8,273 (3,220) | 0.742 (0.731-0.752) | 0.530 (0.513-0.548) | 0.856 (0.813-0.838) | 0.670 (0.648-0.691) | 0.929 (0.924-0.934) | 0.826 (0.815-0.836) |

**Extended Table 3. AI model performance for Gleason score prediction across geographic regions in ProMort I and ProMort II datasets.** Subgroup analysis is presented showing agreements between AI predictions and the pathologist reference standard across different Swedish counties and municipalities by LWK, QWK, and C-index with 95% confidence intervals. Metrics are presented for all cores and for malignant cores only (cores diagnosed as malignant by the reference standard). Core numbers are shown with the number of malignant cores in parentheses. Abbreviation: C-index=concordance index, GS=Gleason score, LWK=linear weighted kappa, QWK=quadratic weighted kappa.

| Sample collection time | Core no. | ISUP LWK | Malignant ISUP LWK | ISUP QWK | Malignant ISUP QWK | ISUP C-index | Malignant ISUP C-index |
|---|---|---|---|---|---|---|---|
| **ProMort I + ProMort II** | | | | | | | |
| 1998-01-13 ~ 2001-08-07 | 2334 (696) | 0.767 (0.746-0.787) | 0.549 (0.508-0.588) | 0.834 (0.809-0.857) | 0.664 (0.609-0.719) | 0.934 (0.924-0.944) | 0.814 (0.786-0.834) |
| 2001-08-07 ~ 2005-03-01 | 2797 (833) | 0.784 (0.766-0.804) | 0.574 (0.536-0.613) | 0.856 (0.837-0.878) | 0.699 (0.655-0.742) | 0.939 (0.931-0.948) | 0.824 (0.804-0.843) |
| 2005-03-01 ~ 2008-09-24 | 1755 (521) | 0.783 (0.757-0.807) | 0.569 (0.519-0.613) | 0.852 (0.822-0.877) | 0.688 (0.622-0.743) | 0.934 (0.923-0.946) | 0.831 (0.806-0.854) |
| 2008-09-24 ~ 2012-04-18 | 1595 (475) | 0.772 (0.745-0.794) | 0.527 (0.473-0.576) | 0.847 (0.816- 0.872) | 0.635 (0.559-0.704) | 0.933 (0.921-0.946) | 0.793 (0.758-0.828) |
| 2012-04-18 ~ 2015-11-12 | 387 (114) | 0.790 (0.738-0.834) | 0.562 (0.455-0.642) | 0.868 (0.810- 0.916) | 0.729 (0.618-0.823) | 0.944 (0.923-0.965) | 0.813 (0.751-0.869) |
| **Overall** | 8,868 (2,639) | 0.777 (0.766-0.787) | 0.557 (0.536-0.578) | 0.848 (0.836-0.860) | 0.677 (0.649-0.702) | 0.936 (0.930-0.941) | 0.816 (0.804-0.828) |
| **ProMort I + ProMort II: Skåne-Malmö** | | | | | | | |
| 1998-01-15 ~ 2003-03-10 | 260 (86) | 0.811 (0.751-0.862) | 0.531 (0.387-0.648) | 0.883 (0.819-0.934) | 0.652 (0.479-0.796) | 0.932 (0.901-0.959) | 0.794 (0.717-0.869) |
| 2003-03-10 ~ 2008-05-03 | 270 (89) | 0.834 (0.780-0.878) | 0.625 (0.505-0.726) | 0.900 (0.840-0.948) | 0.763 (0.601-0.888) | 0.937 (0.904-0.965) | 0.834 (0.741-0.911) |
| 2008-05-03 ~ 2013-06-27 | 241 (81) | 0.802 (0.741-0.861) | 0.605 (0.479-0.713) | 0.863 (0.797-0.925) | 0.733 (0.583-0.856) | 0.938 (0.906-0.963) | 0.811 (0.745-0.874) |
| **Overall** | 771 (256) | 0.816 (0.783-0.845) | 0.587 (0.513-0.649) | 0.882 (0.845-0.912) | 0.717 (0.625-0.795) | 0.935 (0.916-0.952) | 0.814 (0.767-0.856) |

**Extended Table 4. AI model performance for ISUP grade prediction across different sample collection periods.** The temporal robustness of the AI model performance is evaluated with stratified sampling to achieve comparable ISUP grade distributions across time bins. ProMort I and II datasets were merged for analysis. LWK, QWK, and C-index with 95% confidence intervals are shown for agreement between AI predictions and pathologist reference standard. Metrics are presented for all cores and for malignant cores only (cores diagnosed as malignant by the reference standard). Core numbers are shown with the number of malignant cores in parentheses. Abbreviation: C-index=concordance index, ISUP=International Society of Urological Pathology, LWK=linear weighted kappa, QWK=quadratic weighted kappa.

| Sample collection time | Core no. | GS LWK | Malignant GS LWK | GS QWK | Malignant GS QWK | GS C-index | Malignant GS C-index |
|---|---|---|---|---|---|---|---|
| **ProMort I + ProMort II** | | | | | | | |
| 1998-01-13 ~ 2001-08-07 | 2334 (696) | 0.715 (0.692-0.738) | 0.503 (0.461-0.540) | 0.785 (0.755-0.813) | 0.636 (0.588-0.682) | 0.936 (0.926-0.946) | 0.828 (0.801-0.853) |
| 2001-08-07 ~ 2005-03-01 | 2797 (833) | 0.738 (0.717-0.758) | 0.535 (0.498-0.569) | 0.815 (0.790-0.839) | 0.678 (0.640-0.717) | 0.940 (0.931-0.948) | 0.828 (0.818-0.857) |
| 2005-03-01 ~ 2008-09-24 | 1755 (521) | 0.732 (0.707-0.757) | 0.523 (0.482-0.569) | 0.808 (0.776-0.839) | 0.675 (0.628-0.727) | 0.936 (0.925-0.947) | 0.848 (0.824-0.871) |
| 2008-09-24 ~ 2012-04-18 | 1595 (475) | 0.719 (0.691-0.743) | 0.476 (0.427-0.523) | 0.809 (0.777-0.835) | 0.601 (0.541-0.661) | 0.936 (0.923-0.949) | 0.813 (0.781-0.843) |
| 2012-04-18 ~ 2015-11-12 | 387 (114) | 0.732 (0.679-0.778) | 0.491 (0.395-0.588) | 0.817 (0.749-0.870) | 0.648 (0.545-0.745) | 0.946 (0.923-0.964) | 0.826 (0.773-0.877) |
| **Overall** | 8,868 (2,639) | 0.727 (0.715-0.739) | 0.511 (0.489-0.533) | 0.805 (0.790-0.819) | 0.651 (0.625-0.673) | 0.937 (0.932-0.943) | 0.832 (0.820-0.844) |
| **ProMort I + ProMort II: Skåne-Malmö** | | | | | | | |
| 1998-01-15 ~ 2003-03-10 | 260 (86) | 0.762 (0.700-0.822) | 0.483 (0.348-0.595) | 0.843 (0.769-0.906) | 0.622 (0.457-0.745) | 0.931 (0.901-0.959) | 0.796 (0.722-0.868) |
| 2003-03-10 ~ 2008-05-03 | 270 (89) | 0.775 (0.721-0.823) | 0.539 (0.413-0.643) | 0.860 (0.793-0.912) | 0.684 (0.532-0.801) | 0.939 (0.907-0.967) | 0.845 (0.742-0.921) |
| 2008-05-03 ~ 2013-06-27 | 241 (81) | 0.719 (0.658-0.778) | 0.487 (0.378-0.581) | 0.811 (0.747-0.873) | 0.643 (0.525-0.740) | 0.944 (0.912-0.969) | 0.844 (0.783-0.900) |
| **Overall** | 771 (256) | 0.752 (0.718-0.782) | 0.503 (0.438-0.564) | 0.837 (0.795-0.873) | 0.650 (0.575-0.720) | 0.937 (0.918-0.953) | 0.827 (0.780-0.870) |

**Extended Table 5. AI model performance for Gleason score prediction across different sample collection periods.** The temporal robustness of the AI model performance is evaluated with stratified sampling to achieve comparable ISUP grade distributions across time bins. ProMort I and II datasets were merged for analysis. LWK, QWK, and C-index with 95% confidence intervals are shown for agreement between AI predictions and pathologist reference standard. Metrics are presented for all cores and for malignant cores only (cores diagnosed as malignant by the reference standard). Core numbers are shown with the number of malignant cores in parentheses. Abbreviation: C-index=concordance index, GS=Gleason score, LWK=linear weighted kappa, QWK=quadratic weighted kappa.

| Inter-observer | | ISUP LWK | ISUP QWK | ISUP C-index | GS LWK | GS QWK | GS C-index |
|---|---|---|---|---|---|---|---|
| **ProMort I** | | | | | | | |
| Pathologist vs Pathologist | F.G. vs L.M. | 0.240 (0.202-0.278) | 0.526 (0.476-0.572) | 0.771 (0.745-0.797) | 0.249 (0.204-0.290) | 0.544 (0.478-0.597) | 0.771 (0.744-0.795) |
| | F.G. vs M.F. | 0.591 (0.547-0.635) | 0.749 (0.705-0.789) | 0.848 (0.827-0.868) | 0.591 (0.541-0.634) | 0.740 (0.690-0.783) | 0.848 (0.827-0.870) |
| | L.M. vs M.F. | 0.397 (0.351-0.440) | 0.610 (0.561-0.654) | 0.840 (0.812-0.866) | 0.406 (0.347-0.459) | 0.627 (0.559-0.686) | 0.840 (0.812-0.866) |
| AI vs Pathologist | AI vs F.G. | 0.538 (0.489-0.585) | 0.673 (0.613-0.7232) | 0.804 (0.775-0.831) | 0.519 (0.469-0.567) | 0.661 (0.599-0.716) | 0.806 (0.779-0.832) |
| | AI vs L.M. | 0.345 (0.300-0.388) | 0.513 (0.450-0.572) | 0.806 (0.774-0.837) | 0.364 (0.313-0.416) | 0.547 (0.475-0.615) | 0.809 (0.778-0.837) |
| | AI vs M.F. | 0.543 (0.494-0.588) | 0.664 (0.601-0.715) | 0.795 (0.769-0.820) | 0.523 (0.471-0.571) | 0.651 (0.586-0.707) | 0.797 (0.770-0.822) |
| **AI vs Pathologists (average)** | | 0.475 (0.438-0.509) | 0.617 (0.566-0.665) | 0.802 (0.777-0.824) | 0.469 (0.427-0.506) | 0.620 (0.565-0.668) | 0.804 (0.780-0.827) |
| **ProMort II** | | | | | | | |
| Pathologist vs Pathologist | F.G. vs M.F. | 0.905 (0.878-0.929) | 0.961 (0.940-0.976) | 0.953 (0.938-0.966) | 0.886 (0.859-0.911) | 0.953 (0.935-0.968) | 0.951 (0.936-0.965) |
| | F.G. vs Os.A. | 0.820 (0.781-0.856) | 0.909 (0.871-0.943) | 0.941 (0.923-0.957) | 0.783 (0.742-0.818) | 0.886 (0.840-0.922) | 0.938 (0.921-0.955) |
| | M.F. vs Os.A. | 0.798 (0.758-0.836) | 0.890 (0.849-0.927) | 0.935 (0.914-0.952) | 0.763 (0.717-0.802) | 0.865 (0.813-0.905) | 0.932 (0.912-0.952) |
| AI vs Pathologist | AI vs F.G. | 0.848 (0.813-0.878) | 0.930 (0.900-0.954) | 0.935 (0.837-0.917) | 0.794 (0.759-0.829) | 0.899 (0.865-0.926) | 0.942 (0.926-0.956) |
| | AI vs M.F. | 0.833 (0.797-0.867) | 0.912 (0.872-0.943) | 0.926 (0.817- 0.892) | 0.783 (0.747-0.818) | 0.886 (0.849-0.918) | 0.931 (0.914-0.947) |
| | AI vs Os.A. | 0.755 (0.713-0.798) | 0.860 (0.815-0.899) | 0.896 (0.746-0.866) | 0.693 (0.644-0.739) | 0.805 (0.752-0.856) | 0.901 (0.877-0.922) |
| **AI vs Pathologists (average)** | | 0.812 (0.779-0.841) | 0.901 (0.870-0.926) | 0.919 (0.901-0.934) | 0.757 (0.719-0.787) | 0.864 (0.830-0.892) | 0.925 (0.908-0.939) |

**Extended Table 6. Inter-observer agreement for ISUP grade and Gleason score.** Pairwise concordance between individual observers (pathologists and AI model) compared to all other observers. LWK, QWK, and C-index with 95% confidence intervals are shown for ISUP grade and Gleason score. ProMort I subset includes 548 cores annotated by F.G., L.M., and M.F., ProMort II subset includes 347 cores annotated by F.G., M.F., and Os.A. Abbreviation: C-index=concordance index, GS=Gleason score, ISUP=International Society of Urological Pathology, LWK=linear weighted kappa, QWK=quadratic weighted kappa.

| ISUP grade | Gleason score | n | Events | Median FU (yr) | HR | 95% CI | p-value |
|---|---|---|---|---|---|---|---|
| 0 | 0+0 | 58 | 18 | 8.1 | 1.22 | 0.72-2.07 | 0.460 |
| 1 | 3+3 | 209 | 26 | 9.0 | Reference | | |
| 2 | 3+4 | 198 | 71 | 9.2 | 1.27 | 0.90-1.80 | 0.176 |
| 3 | 4+3 | 73 | 40 | 7.1 | 2.51 | 1.68-3.76 | $7.4\times10^{-6}$ |
| 4 | 3+5 | 2 | 2 | 10.0 | 3.23 | 0.79-13.23 | 0.103 |
| | 4+4 | 104 | 78 | 7.1 | 3.68 | 2.62-5.17 | $6.5\times10^{-14}$ |
| 5 | 4+5 | 257 | 198 | 5.3 | 4.72 | 3.52-6.34 | $4.9\times10^{-25}$ |
| | 5+4 | 28 | 28 | 5.4 | 6.32 | 4.02-9.94 | $1.5\times10^{-15}$ |
| | 5+5 | 36 | 30 | 3.6 | 6.84 | 4.40-10.65 | $1.6\times10^{-17}$ |

**Extended Table 7. Hazard ratios for prostate cancer–specific mortality by AI-predicted grade.** Cox proportional hazards regression estimates with ISUP grade 1 (Gleason score 3+3) as reference category. For ISUP grades 4 and 5, which encompass multiple Gleason scores, hazard ratios are additionally reported at the Gleason score level. Gleason score 3+5 (n=2) is included for completeness but should be interpreted with caution due to the small sample size. Abbreviation: CI=confidence interval, FU=follow-up, HR=hazard ratio, ISUP=International Society of Urological Pathology, n=number of patients, yr=year.

| Source | Core no. | Sensitivity | | Specificity | | GS QWK | | GS C-index | |
|---|---|---|---|---|---|---|---|---|---|
| | | UNI | Virchow2 | UNI | Virchow2 | UNI | Virchow2 | UNI | Virchow2 |
| **ProMort I** | | | | | | | | | |
| Skåne-Helsingborg | 462 (186) | 1.000 (1.000-1.000) | 1.000 (1.000-1.000) | 0.895 (0.859-0.931) | 0.891 (0.853-0.928) | 0.805 (0.735-0.866) | 0.819 (0.763-0.863) | 0.936 (0.920-0.950) | 0.930 (0.913-0.946) |
| Skåne-Kristianstad | 355 (201) | 0.990 (0.975-1.000) | 1.000 (1.000-1.000) | 0.922 (0.880-0.962) | 0.890 (0.839-0.938) | 0.849 (0.786-0.892) | 0.800 (0.727-0.858) | 0.936 (0.918-0.952) | 0.923 (0.901-0.941) |
| Skåne-Lund | 542 (231) | 0.983 (0.962-0.996) | 0.991 (0.978-1.000) | 0.830 (0.783-0.872) | 0.859 (0.822-0.898) | 0.729 (0.654-0.798) | 0.755 (0.686-0.819) | 0.917 (0.900-0.934) | 0.929 (0.901-0.941) |
| Skåne-Malmö | 572 (232) | 0.983 (0.964-0.996) | 0.974 (0.953-0.991) | 0.853 (0.817-0.888) | 0.891 (0.859-0.924) | 0.776 (0.713-0.831) | 0.771 (0.706-0.824) | 0.926 (0.908-0.940) | 0.934 (0.914-0.944) |
| Örebro | 162 (63) | 0.984 (0.948-1.000) | 0.984 (0.941-1.000) | 0.899 (0.837-0.957) | 0.889 (0.822-0.949) | 0.770 (0.666-0.857) | 0.701 (0.534-0.834) | 0.930 (0.900-0.955) | 0.913 (0.918-0.948) |
| **Overall** | 2,093 (913) | 0.988 (0.981-0.995) | 0.990 (0.983-0.996) | 0.870 (0.849-0.889) | 0.882 (0.864-0.900) | 0.785 (0.751-0.815) | 0.780 (0.748-0.811) | 0.925 (0.917-0.933) | 0.928 (0.921-0.936) |
| **ProMort II** | | | | | | | | | |
| Dalarna | 922 (400) | 0.993 (0.983-1.000) | 0.995 (0.987-1.000) | 0.868 (0.839-0.895) | 0.899 (0.873-0.924) | 0.855 (0.819-0.888) | 0.833 (0.798-0.868) | 0.940 (0.929-0.952) | 0.934 (0.921-0.947) |
| Gävleborg | 751 (349) | 0.991 (0.980-1.000) | 0.986 (0.972-0.997) | 0.876 (0.844-0.907) | 0.903 (0.873-0.930) | 0.864 (0.831-0.893) | 0.845 (0.814-0.873) | 0.935 (0.921-0.947) | 0.927 (0.912-0.939) |
| Halland | 766 (341) | 0.991 (0.981-1.000) | 0.997 (0.991-1.000) | 0.889 (0.860-0.919) | 0.897 (0.864-0.927) | 0.799 (0.750-0.847) | 0.803 (0.758-0.844) | 0.926 (0.910-0.943) | 0.929 (0.913-0.944) |
| Jönköping | 709 (238) | 0.996 (0.987-1.000) | 1.000 (1.000-1.000) | 0.822 (0.789-0.854) | 0.871 (0.841-0.902) | 0.786 (0.736-0.835) | 0.778 (0.729-0.826) | 0.932 (0.916-0.946) | 0.937 (0.923-0.951) |
| Kalmar | 482 (155) | 0.994 (0.976-1.000) | 0.994 (0.978-1.000) | 0.881 (0.845-0.913) | 0.914 (0.885-0.944) | 0.752 (0.682-0.822) | 0.802 (0.738-0.860) | 0.931 (0.911-0.949) | 0.946 (0.929-0.960) |
| Kronoberg | 261 (95) | 0.990 (0.966-1.000) | 0.990 (0.966-1.000) | 0.904 (0.855-0.947) | 0.898 (0.850-0.942) | 0.863 (0.780-0.926) | 0.814 (0.733-0.888) | 0.956 (0.930-0.975) | 0.940 (0.916-0.962) |
| Norrbotten | 531 (211) | 0.995 (0.984-1.000) | 0.995 (0.985-1.000) | 0.850 (0.812-0.887) | 0.913 (0.880-0.943) | 0.807 (0.750-0.856) | 0.793 (0.741-0.840) | 0.928 (0.910-0.945) | 0.934 (0.918-0.950) |
| Skåne-Helsingborg | 587 (197) | 1.000 (1.000-1.000) | 1.000 (1.000-1.000) | 0.864 (0.931-0.897) | 0.885 (0.852-0.916) | 0.685 (0.597-0.766) | 0.781 (0.715-0.841) | 0.927 (0.908-0.945) | 0.943 (0.928-0.958) |
| Skåne-Kristianstad | 496 (204) | 0.97 (0.953-0.995) | 0.995 (0.985-1.000) | 0.829 (0.784-0.870) | 0.846 (0.807-0.885) | 0.729 (0.668-0.788) | 0.784 (0.724-0.838) | 0.887 (0.859-0.912) | 0.917 (0.895-0.936) |
| Skåne-Lund | 557 (140) | 0.971 (0.941-0.993) | 0.993 (0.975-1.000) | 0.820 (0.783-0.858) | 0.851 (0.816-0.885) | 0.673 (0.599-0.741) | 0.687 (0.612-0.755) | 0.912 (0.886-0.935) | 0.925 (0.904-0.945) |
| Skåne-Malmö | 733 (203) | 0.995 (0.984-1.000) | 0.990 (0.975-1.000) | 0.889 (0.863-0.915) | 0.911 (0.886-0.934) | 0.755 (0.694-0.811) | 0.768 (0.716-0.816) | 0.941 (0.927-0.954) | 0.946 (0.933-0.957) |
| Värmland | 594 (331) | 0.994 (0.984-1.000) | 1.000 (1.000-1.000) | 0.924 (0.888-0.954) | 0.924 (0.894-0.954) | 0.884 (0.849-0.914) | 0.891 (0.865-0.915) | 0.941 (0.926-0.956) | 0.948 (0.936-0.959) |
| Västmanland | 548 (177) | 0.989 (0.971-1.000) | 1.000 (1.000-1.000) | 0.911 (0.881-0.939) | 0.916 (0.890-0.944) | 0.8446 (0.788-0.897) | 0.799 (0.737-0.856) | 0.952 (0.936-0.966) | 0.946 (0.930-0.961) |
| Örebro | 336 (179) | 1.000 (1.000-1.000) | 0.989 (0.971-1.000) | 0.854 (0.797-0.909) | 0.892 (0.838-0.938) | 0.777 (0.706-0.843) | 0.763 (0.702-0.823) | 0.907 (0.878-0.931) | 0.908 (0.879-0.934) |
| **Overall** | 8,273 (3,220) | 0.992 (0.988-0.995) | 0.995 (0.992-0.997) | 0.868 (0.859-0.878) | 0.894 (0.885-0.902) | 0.803 (0.790-0.817) | 0.807 (0.794-0.820) | 0.930 (0.926-0.935) | 0.935 (0.931-0.939) |

**Extended Table 8. Performance of foundation models UNI and Virchow2 for prostate cancer detection and grading on ProMort I and ProMort II.** Results are presented showing agreements between model predictions and the pathologist reference standard across different Swedish counties and municipalities by sensitivity and specificity for cancer detection, QWK and C-index for Gleason score grading, with 95% confidence intervals. Core numbers are shown with the number of malignant cores in parentheses. Abbreviation: C-index=concordance index, GS=Gleason score, QWK=quadratic weighted kappa.

| No. | STARD-AI Item | Reported on page/section | Context/Evidence |
|---|---|---|---|
| **Title or abstract** | | | |
| 1 | Identification as a study reporting AI-centred diagnostic accuracy and reporting at least one measure of accuracy within title or abstract | Abstract | "Validation of an AI-based end-to-end model for prostate pathology", QWK reported in abstract |
| **Abstract** | | | |
| 2 | Structured summary of study design, methods, results and conclusions (for specific guidance, please see STARD for Abstracts) | Abstract | Abstract covers design, cohort, primary metric, key conclusions |
| **Introduction** | | | |
| 3 | Scientific and clinical background, including the intended use of the index test, whether it is novel or an established index test, and its integration into an existing or new workflow, if applicable | Pages 1–2 (Introduction); Study protocol (Introduction) | Motivation for Gleason grading AI, GleasonAI description, clinical deployment context |
| 4 | Study objectives and hypotheses | Page 2 (Introduction); Study protocol (Methods and Analysis: Study Objectives) | Evaluating GleasonAI generalizability across geography, inter-observer, and temporal variation |
| **Methods** | | | |
| 5 | Whether data collection was planned before the index test and reference standard were performed (prospective study) or after (retrospective study) | Page 2 (Introduction); Study protocol (Study design) | Retrospective; diagnostic specimens from 1998–2015 archival material, digitization and reference standard 2015-2018 |
| 6 | Formal approval from an ethics committee. If not required, justify why | Page 15 (Ethics declarations); Study protocol (Ethics and dissemination) | Permits 2012/1586–31/1, 2016/613–31/2, 2019–01395, 2019–05220 |
| 7 | Eligibility criteria: listing separate inclusion and exclusion criteria in the order that they are applied at both participant level and data level | Pages 9–11 (Methods: Study materials); Study protocol (Data sources; Preanalysis exclusion criteria; Figure 1, Figure 2) | Original ProMort I and ProMort II inclusion criteria, re-grading samples inclusion criteria, slide quality exclusions, segmentation failures, annotation errors |
| 8 | On what basis potentially eligible participants were identified (such as symptoms, results from previous tests, inclusion in registry) | Pages 9–11 (Methods: Study materials); Study protocol (Data sources) | Derived from NPCR; case-control selection based on prostate cancer death and matched controls |
| 9 | Where and when potentially eligible participants were identified (setting, location, and dates) | Pages 9–11 (Methods: Study materials); Study protocol (Data sources) | Sweden; ProMort I: 1998–2011 diagnoses; ProMort II: 1998–2015 diagnoses; digitized 2015–2018 |
| 10 | Whether participants formed a consecutive, random, or convenience series | Pages 9–11 (Methods: Study materials); Study protocol (Data sources) | Nested case-control; matched 1:1 on year and county/institution; digitization from selected counties for ProMort I and all retrievable samples for ProMort II |
| 11 | Source of the data and whether it has been routinely collected, specifically collected for the purpose of the study or acquired from an open-source repository | Pages 9–11 (Methods: Study materials); Study protocol (Data sources, regarding protocols in supplemental materials) | Routine diagnostic archival material from NPCR; digitised specifically for ProMort studies |
| 12 | Who undertook the annotations for the dataset (including experience levels and background) and how (within the same clinical context or in a post-hoc fashion), if applicable | Pages 10–11 (Methods: Study materials); Study protocol (Reference standard protocols, Tables 3–4) | Four genitourinary pathologists (F.G., L.M., M.F., Os.A.) with 2–21 years experience; independent blinded review via virtual microscopy |
| 13 | Devices (manufacturer, model) that were used to capture data; software (with version number) used to engineer the index test, highlighting the intended use | Pages 10–11 (Methods: Study materials); Page 14 (Computing hardware); Study protocol (Slide digitisation sections) | 3DHistech Pannoramic 250 Flash II scanner at 40×; Python 3.8.10, PyTorch 2.0.0, timm 0.9.8, etc. |
| 14 | Data acquisition protocols (e.g. contrast protocol or reconstruction method for medical images) and details of data pre-processing in sufficient detail to allow replication | Page 11 (Methods: WSI pre-processing); Study protocol (AI system) | 512×512px patches at 8.0 μm/px for segmentation; 256×256px at 1.0 μm/px for classification; ≥10% tissue per patch |

| 15a | Index test, in sufficient detail to allow replication | Pages 11–12 (Methods: Deep learning models); Study protocol (AI system) | ABMIL with EfficientNet-V2-S encoder; gated attention aggregator; ensemble of 10 models with deterministic test-time augmentation |
|---|---|---|---|
| 15b | How the index test was developed, including any training, validation, testing and external evaluation, detailing sample sizes, when applicable | Page 12 (Methods: Model training); Study protocol (Model development data) | Trained on 61,483 WSIs from 4,467 patients across 3 clinical cohorts; 10-fold cross-validation ensemble |
| 15c | Definition of and rationale for test positivity cut-offs or result categories of the index test, distinguishing pre-specified from exploratory | Page 12 (Methods: Model prediction); Study protocol (System output) | ISUP grade derived from majority voting of 10 models; cancer positive when ISUP grade > 0; cancer probability = 1 minus min benign probability of Gleason pattern 1 or 2, no cut-off manually chosen |
| 15d | The specified end user of the index test and the level of expertise required of users | Study protocol (Study objectives) | Clinical implementation and user interaction explicitly stated as outside scope; model designed as pathologist decision-support |
| 16a | Reference standard, in sufficient detail to allow replication | Pages 10–11 (Methods: Study materials); Study protocol (Reference standard protocols, Tables 3–4) | Core-level Gleason grading by F.G. (primary) following 2014 ISUP guidelines; spatial delineation of each core via virtual microscopy platform |
| 16b | Rationale for choosing the reference standard (if alternatives exist) | Page 3 (Results: Dataset characteristics); Study protocol (Definition of reference standard) | F.G. selected as primary reference for consistent coverage across majority of samples (99.9% ProMort I, 96.1% ProMort II) |
| 16c | Definition of and rationale for test positivity cut-offs or result categories of the reference standard, distinguishing pre-specified from exploratory | Pages 10–11 (Methods: Study materials); Study protocol (Core-level outcomes, Definition of reference standard) | 2014 ISUP grading guidelines; high-grade pattern (Gleason 4 or 5) designated as secondary regardless of extent; benign = GS 0+0 |
| 17a | Whether clinical information and reference standard results were available to the performers or readers of the index test | Study protocol (Study design) | No, AI model processed images without access to reference standard; fully blinded evaluation |
| 17b | Whether clinical information and index test results were available to the assessors of the reference standard | Pages 10–11 (Methods: Study materials); Study protocol (Reference standard protocols) | Pathologists were blinded to original clinical/histopathological information and case-control status |
| 18 | Methods for estimating or comparing measures of diagnostic accuracy | Page 13 (Method: Statistical analysis); Study protocol (Statistical analyses; Table 5) | QWK, LWK, C-index, sensitivity/specificity, AUROC, calibration curve; bootstrap CIs (n=1000); Cox regression for survival |
| 19 | How indeterminate index test or reference standard results were handled | Page 3 (Results: Dataset characteristics); Page 13 (Method: Statistical analysis); Study protocol (Preanalysis exclusion criteria) | Segmentation failures reported separately and excluded; annotation errors identified and excluded; no indeterminate category for classification |
| 20 | How missing data on the index test and reference standard were handled | Page 3 (Results); Study protocol (Preanalysis exclusion criteria, Figure 1 –2) | Cores without reference annotation excluded; three segmentation failures excluded and documented; slides lacking H&E excluded pre-analysis |
| 21 | Any analyses of variability in diagnostic accuracy, distinguishing pre-specified from exploratory | Pages 4–7 (Results); Study protocol (Statistical analyses overview) | Pre-specified: geographic subgroup, temporal sensitivity, inter-observer analysis; exploratory: foundation model comparison, survival analysis |
| 22 | Intended sample size and how it was determined | Pages 9–10 (Methods: Study materials); Study protocol (Data sources, Figures 1–2) | Sample size determined by availability of ProMort case-control cohorts; 10,366 cores from 1,028 patients after quality exclusions |
| 23 | Details of any performance error analysis, and algorithmic bias and fairness assessments if undertaken | Pages 12–13 (Methods: Statistical analysis); Study protocol (Statistical analyses, Table 5) | Report pre-processing failures in the AI-based tissue segmentation process; AUROC and C-index; temporal subgroup analysis |
| **Results** | | | |
| 24 | Flow of participants, using a diagram | Study protocol (Figure 1–2) | CONSORT-style flow charts for ProMort I and ProMort II inclusion/exclusion |
| 25 | Baseline demographic, clinical and technical characteristics of training, validation and test set, if applicable | Figure 1; Study protocol (Tables 1–2) | Geographic distribution, cancer grade distribution; patient-level clinical characteristics |
| 26a | Distribution of severity of disease in those with the target condition | Figure 1b, Extended Table 1 | Core-level ISUP grade and Gleason score distribution by dataset and geographic region |
| 26b | Distribution of alternative diagnoses in those without the target condition | Figure 1b; Extended Table 1 | Proportion of benign cores per region shown (ISUP 0: ~49.5% ProMort I, ~53.4% ProMort II) |
| 27 | Time interval and any clinical interventions between index test and reference standard | Pages 9–10 (Methods: Study materials); Study protocol (Data sources) | Diagnostic slides from 1998–2015; digitized 2015–2018; re-annotated 2017–2020 by expert pathologists, no clinical interventions between |
| 28 | Whether the datasets represent the distribution of the target condition that one would expect from the intended use population | Pages 4 (Results: Dataset characteristics); Study protocol (Generalisability limitations) | Case-enriched design (50% mortality) noted as a limitation; grade distribution consistent with cohort selection criteria |
| 29 | For external evaluation on an independent dataset, an assessment of how this differs from the training, validation and test sets | Pages 2 (Introduction); Study protocol (Study design, Data independence) | Different scanner (3DHistech vs Aperio/Hamamatsu/Philips), different laboratories, different geographic population; no patient overlap |
| 30 | Cross tabulation of the index test results (or their distribution) by the results of the reference standard | Figure 3a; Extended Figure 3; Extended Tables 2–3 | Confusion matrices for ISUP grading (Figure 3a) and cancer detection (Extended Figure 3) per region |
| 31 | Estimates of diagnostic accuracy and their precision (such as 95% confidence intervals) | Pages 3–6 (Results); Extended Tables 2–8 | QWK, LWK, C-index, sensitivity, specificity, AUC all reported with 95% bootstrap CIs |
| 32 | Any adverse events from performing the index test or the reference standard | Not applicable (retrospective digital pathology study) | No patient-level adverse events; retrospective digital study on archival slides poses no clinical risk |
| **Discussion** | | | |
| 33 | Study limitations, including sources of potential bias, statistical uncertainty, and generalisability | Pages 8–9 (Discussion); Study protocol (Limitations and interpretive considerations) | Swedish-only cohort, patch extraction within annotated areas only, heterogeneous annotation granularity; spectrum and prevalence bias discussed |
| 34 | Implications for practice, including the intended use and clinical role of the index test | Pages 7–9 (Discussion) | Supports digitized archive repurposing, consistent second-opinion grading; temporal robustness enables retrospective prognostic research |
| 35 | Ethical considerations and adherence to ethical standards associated with the use of the index test and issues of fairness | Page 15 (Ethics declarations); Study protocol (Ethics and dissemination) | Declaration of Helsinki; GDPR; Swedish Patient Data Act; approved ethics permits listed; geographic fairness addressed via subgroup analysis |
| **Other information** | | | |
| 36 | Registration number and name of registry | Not formally registered; Study protocol (Study status) | No formal registry; pre-specified protocol posted on medRxiv and published in BMJ Open prior to validation execution, serving equivalent purpose |
| 37 | Where the full study protocol can be accessed | Page 2 (Introduction, reference 16); Study protocol (full document) | Study protocol published in BMJ Open (Ji et al., BMJ Open 2025;15:e111361) |
| 38 | Sources of funding and other support; role of funders | Page 15 (Acknowledgements) | SciLifeLab/Wallenberg, Swedish Cancer Society, NAISS/SNIC, Berzelius; funder did not influence results |
| 39 | Commercial interests, if applicable | Page 14 (Competing interests) | N.M., K.K., M.E. are shareholders/co-founders of Clinsight AB; other authors declare no conflicts |
| 40a | Availability of datasets and code; detailing any restrictions on their reuse and repurposing | Page 14 (Data availability; Code availability) | Image data available from corresponding author on reasonable request; outcome data not shared; no custom code, analysis scripts available on request |
| 40b | Whether outputs are stored, auditable and available for evaluation, if necessary | Study protocol (Study design, Study status) | Pre-specified protocol publicly registered before validation; model locked before evaluation; auditable workflow described in study status section |

**Extended Table 9. STARD-AI checklist.**